\RequirePackage[l2tabu]{nag}
\documentclass{article}

\usepackage[utf8]{inputenc}
\usepackage[T1]{fontenc}

\usepackage{geometry}
\usepackage{color}
\usepackage{color,soul}
\usepackage{amsmath}

\usepackage{comment} 
\excludecomment{manuscriptComment} 
\usepackage{cite}

\usepackage{lmodern}
\usepackage[tracking=true,kerning=true,final]{microtype}

\usepackage{hyperref}
\hypersetup{colorlinks=true, linkcolor=blue,  anchorcolor=blue, citecolor=blue, filecolor=blue, menucolor=blue, urlcolor=blue}
\hypersetup{pdfauthor={},pdftitle={}}

\usepackage{natbib}

\usepackage[caption=false]{subfig} 
\usepackage{caption}   
\usepackage{color}
\usepackage{graphicx}
\graphicspath{{./}{./figs/}}
\usepackage{booktabs} 

\usepackage{cite}
\usepackage{natbib} 

\usepackage{amsfonts}
\usepackage{bm}
\usepackage{mathtools}
\usepackage{amsthm}
\usepackage{mathrsfs}

\usepackage{algorithm}
\usepackage{algpseudocode}

\usepackage{amsmath}
\usepackage{amssymb}

\usepackage{authblk}
\usepackage{comment}

\title{Multifidelity digital twin for real-time monitoring of structural dynamics in aquaculture net cages}

\author[1, *]{Eirini Katsidoniotaki}
\author[2]{Biao Su}
\author[2]{Eleni Kelasidi}
\author[1, *]{Themistoklis P. Sapsis}

\affil[1]{Department of Mechanical Engineering, Massachusetts Institute of Technology, Cambridge, MA 02138, USA}
\affil[2]{Department of Aquaculture Technology, SINTEF Ocean, Trondheim, Norway}

\affil[*]{eirka289@mit.edu}
\affil[*]{sapsis@mit.edu}

\date{\today}

\usepackage{amsmath}
\usepackage{amssymb}

\usepackage[normalem]{ulem}

\begin{document}

\maketitle\

\begin{abstract}
As the global population grows and climate change intensifies, sustainable food production is critical. Marine aquaculture offers a viable solution, providing a sustainable protein source. However, the industry's expansion requires novel technologies for remote management and autonomous operations. Digital twin technology can advance the aquaculture industry, but its adoption has been limited. Fish net cages, which are flexible floating structures, are critical yet vulnerable components of aquaculture farms. Exposed to harsh and dynamic marine environments, the cages experience significant loads and risk damage, leading to fish escapes, environmental impacts, and financial losses. We propose a multifidelity surrogate modeling framework for integration into a digital twin for real-time monitoring of aquaculture net cage structural dynamics under stochastic marine conditions. Central to this framework is the nonlinear autoregressive Gaussian process method, which learns complex, nonlinear cross-correlations between models of varying fidelity. It combines low-fidelity simulation data with a small set of high-fidelity field sensor measurements, which offer the real dynamics but are costly and spatially sparse. Validated at the SINTEF ACE fish farm in Norway, our digital twin receives online metocean data and accurately predicts net cage displacements and mooring line loads, aligning closely with field measurements. The proposed framework is beneficial where application-specific data are scarce, offering rapid predictions and real-time system representation. The developed digital twin prevents potential damages by assessing structural integrity and facilitates remote operations with unmanned underwater vehicles. Our work also compares Gaussian processes and graph convolutional networks for predicting net cage deformation, highlighting the latter's effectiveness in complex structural applications.

\vspace{1cm}
\end{abstract}

\section*{Introduction}

As the global population grows and the effects of climate change escalate, concerns about future food security and human nutrition increase, creating an urgent demand to enhance food production \cite{FP, UN, WEF}. Fish and fishery products have become vital in diets worldwide, particularly in developing countries \cite{han2022drives}, but unsustainable fishing practices threaten natural fish populations \cite{WWF}. Marine aquaculture emerges as a vital solution \cite{WEF}, offering a sustainable protein source that meets the rising global food demands while serving as a response to climate change. Fish farms have lower carbon emissions \cite{gephart2021environmental} compared to other animal proteins and support a diverse, resilient food system. Sustainable growth of aquaculture requires further reducing the carbon footprint of fish farms through strategic farm placement, efficient feed use, minimizing food waste and fish mortality rates, and the adoption of novel technologies to enhance autonomy in challenging fish farm operations. As the industry matures and operations move offshore to harsh and uncertain environments \cite{bjelland2015exposed}, advancing real-time monitoring, operational planning, and decision-making practices is essential for ensuring efficiency, safety, cost-effectiveness, and sustainability \cite{ubina2023digital, fore2024digital}.\\

\noindent Digital twins have great success in various sectors, yet its adoption in aquaculture has been limited, partly because it is a comparatively young industry \cite{fore2024digital}. Digital twin technology can significantly advance aquaculture by enabling fish farm remote monitoring, autonomous operations, management, and predictive analytics \cite{ubina2023digital, fore2024digital}. Integrating real-time observations from in-situ monitoring equipment alongside realistic models within a digital twin framework can provide more detailed and accurate insights into the dynamics of fish farms. Implementing a digital twin typically involves several components \cite{ubina2023digital, fore2024digital}, such as assessing fish health and behavior \cite{saad2023web}, tracking environmental conditions \cite{huan2020prediction}, and monitoring the structural dynamics of fish net cages \cite{su2023towards}. Our study primarily focuses on the latter component.\\

\noindent Fish net cages are flexible floating structures anchored to the seabed, serving as critical yet vulnerable components of aquaculture farms. Exposed to the harsh and dynamic marine environment, these cages experience significant loads and are at risk of damage that can lead to fish escapes, resulting in negative environmental impacts and substantial financial losses \cite{fore2021causal}. Influenced by waves and currents, the net cages are prone to large motions and deformations \cite{su2021integrated} due to their elastic structure.  These deformations reduce the available space for fish, causing decreased oxygen levels and increased stress \cite{turnbull2005stocking}, which negatively impacts fish growth and survival, leading to higher mortality rates and significant food waste. The development of a digital twin capable to real-time monitor the net cage dynamics can serve as an early warning system for fish farm operators, helping to prevent significant reductions in cage volume. Additionally, a digital twin is essential for assessing the structural integrity and preventing potential catastrophic damages, while it can be used to facilitate the autonomous inspection and repair operations with unmanned underwater vehicles (UUVs) in fish farms. For example, when autonomously controlling a remotely operated vehicle (ROV) within the fish cage to assess potential damage, prior knowledge of the cage's shape is crucial for effective path planning and collision avoidance with the net \cite{amundsen2024aquaculture}. \\

\noindent Unlike more rigid marine structures such as floating wind turbine foundations or vessels, the elastic and complex geometry of aquaculture net cages—composed of millions of slender twines—poses unique challenges in predicting their dynamics \cite{martin2021numerical}. Modeling these net cages is a multifaceted task that involves solving sub-problems: one focuses on the interaction between the fluid and a rigid moored-floating structure, and the other addresses the flexible porous net. Accurately depicting the deformation of these cages requires detailed information on the displacements of each twine, necessitating extensive data and computational effort. While high-fidelity numerical models like CFD simulations provide accurate estimations of structural dynamics \cite{martin2021numerical}, they demand vast computational resources, making them unsuitable for real-time forecasting or rapid hindcasting tasks in digital twins. Lower fidelity models \cite{su2019multipurpose, reite2014fhsim, kristiansen2012modelling, huang2006dynamical} are more accessible and less computationally demanding but often fail to capture the real system accurately, particularly under non-linear effects from extreme waves and fluid-structure interactions, thus limiting their integration into digital twins. Conversely, field sensor measurements offer the most accurate depiction of net cage dynamics but are constrained by high sensor costs and the risk of data transmission loss \cite{su2023towards}. Additionally, sensors typically provide localized information, resulting in significant spatial gaps or high sparsity. These challenges underscore the need for innovative approaches to effectively monitor net cage system dynamics in real-time.\\

\noindent Current research is focused on innovative models for precise real-time monitoring of net cage dynamics that can be integrated into digital twins. Su et al. \cite{su2021integrated} introduced a real-time numerical simulation model that uses in-situ, real-time positioning sensor data to dynamically adjust model inputs—specifically, the magnitude and direction of the current—based on the difference between simulated and actual net positions. Building on this, Su et al. further developed their model into a physics-based digital twin \cite{su2023towards} that incorporates in-situ sensor data to accurately simulate real-time cage responses, net deformations, and mooring loads. Although successfully tested at a full-scale aquaculture site, this model faces challenges such as high computational costs and potential inaccuracies due to inconsistent or faulty sensor data caused by harsh environmental conditions, hardware limitations, and unexpected interference, making it less suited for real-time applications. These issues highlight the need for digital twins using data-driven and machine learning (ML)-based surrogate models \cite{liu2023modelling,chakraborty2021role}.\\

\noindent ML-based surrogate models for predicting the dynamic behavior of fish net cages remain an under-explored research area. Some studies utilized artificial neural networks (ANNs) for net damage detection \cite{bi2020efficient, zhao2022digital, zhang2022netting}, yet research on predicting the structural dynamics of aquaculture net cages remains sparse. Zhao et al. \cite{zhao2019prediction} developed a backpropagation ANN with one hidden layer, using training data from low-fidelity numerical simulations, to correlate ocean waves with net cage structural responses, including maximum tension in mooring lines, minimum effective volume ratio of the cage, and maximum stress on the floating collar. Although their model outputs the volume ratio, it does not capture the precise topology of the deformed net under wave excitation. To the best of our knowledge, no surrogate models have been developed to predict the entire net cage deformation, which can be beneficial for preventing catastrophic events and supporting autonomous fish farm operations. This underscores the need for more advanced ML-based surrogate models, built on high-fidelity data, that can accurately predict the complex dynamics of flexible net cages.  To support effective real-time monitoring, these models should be capable of integrating into on-the-fly digital twins, which means their computational demands need to be limited.\\

\noindent Creating surrogate models that accurately represent complex systems often encounters the challenge of acquiring high-fidelity, application-specific data. Multifidelity data assimilation \cite{popov2022multifidelity} involves methods that merge information about the same underlying truth obtained through multiple models or observation operators at different fidelity levels, resulting in a model that best represents the system’s state. Among these methods, optimal interpolation is a widely used data assimilation technique; however, it may exhibit large discrepancies with the truth \cite{babaee2020multifidelity}. To this end, multifidelity surrogate modeling \cite{conti2024multi} aims to leverage machine learning methods with computationally inexpensive low-fidelity data and scarce high-fidelity data to accurately predict the quantities of interest. Despite their limitations, low-fidelity data can reveal system trends and patterns, providing valuable information to complement the limited high-fidelity data used for model training. Multifidelity methods \cite{kennedy2000predicting, perdikaris2017nonlinear, cutajar2019deep, meng2020composite, meng2021multi, zhang2022multi, conti2024multi} for constructing surrogate models aim to achieve effective data fusion from various fidelity levels, enabling strong generalization performance of data-driven models in regions where high-fidelity data are limited. This approach is particularly beneficial for various digital twin applications \cite{champenois2024machine, lai2023digital, desai2023enhanced, chetan2021multi}, enabling accurate, real-time system representation.

\begin{figure}[ht]
\centering
\includegraphics[width=0.85\linewidth]{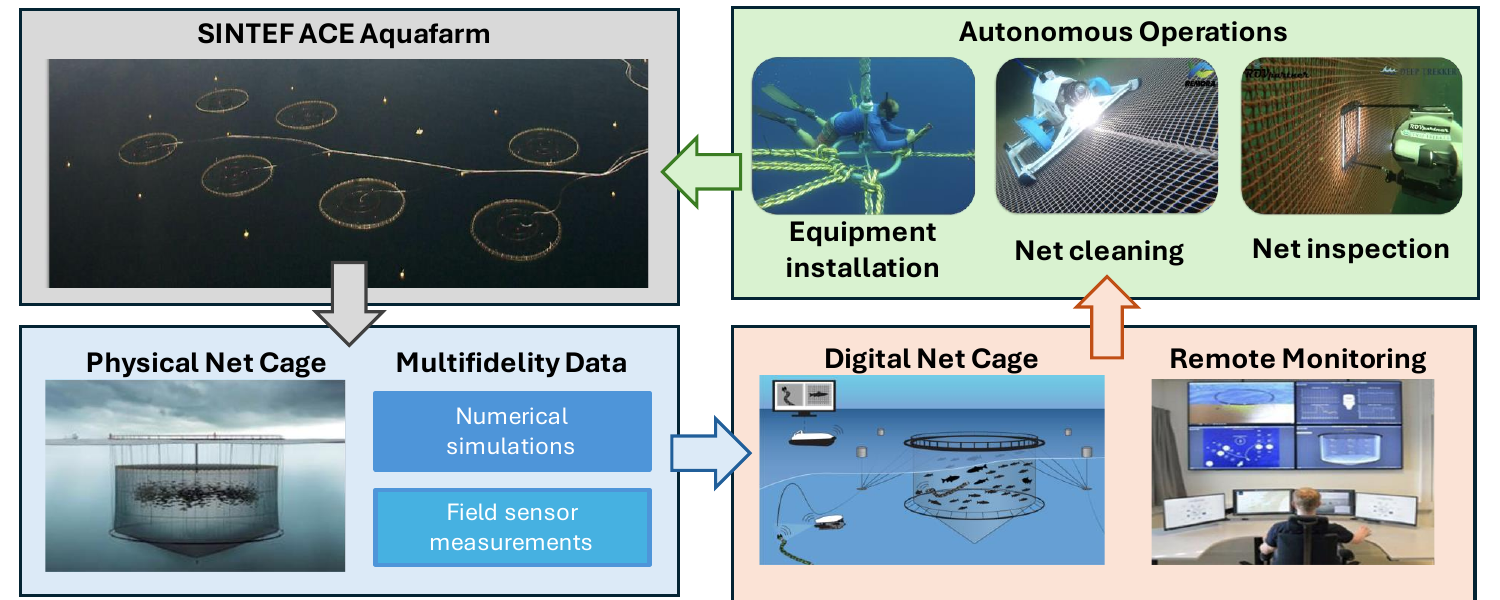}
\includegraphics[width=0.85\linewidth]{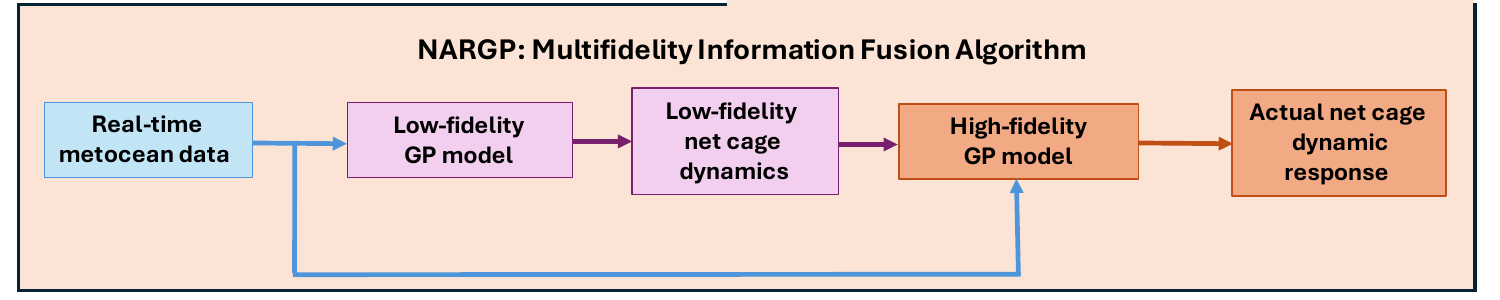}
\caption{\textbf{Real-time monitoring and remote fish farm management using digital twin technology.} The structural response of a net cage to dynamic marine environments is modeled using data from varying fidelity sources, including low-fidelity numerical simulations and high-fidelity field sensor measurements. A digital twin of the physical net cage is developed using the proposed multifidelity framework, central to which is the Nonlinear Autoregressive Gaussian Process (NARGP) method \cite{perdikaris2017nonlinear}. 
}
\label{fig:DT-overview}
\end{figure}

\noindent We propose a framework specifically designed for seamless integration into digital twins, enabling real-time monitoring and visualization of aquaculture net cage structural dynamics under stochastic marine conditions. The framework is built on a multifidelity surrogate modeling method that integrates data from various fidelity sources representing the fish farm environment. Specifically, the framework utilizes the Nonlinear Autoregressive Gaussian Process (NARGP) method \cite{perdikaris2017nonlinear}, which is capable of learning complex, nonlinear, and space-dependent cross-correlations between models of varying fidelity. Due to the reduced algorithmic complexity and computational cost of the NARGP, this framework offers rapid predictions essential for real-time representation of the current system's state. The resulting multifidelity digital twin enables effective monitoring of structural integrity and the detection of irregularities, functioning as an alarm system to guide necessary actions and prevent damage that could lead to fish escapes or mortalities. Moreover, it informs decision-making and enhances operational planning by supporting the control of autonomous underwater operations in fish farms, such as net damage inspections and net cleaning tasks with ROVs. This framework significantly enhances efficiency, automation, safety, and cost savings by reducing the need for additional sensors and personnel visits, which are particularly challenging in harsh marine environments. Figure \ref{fig:DT-overview} illustrates the proposed multifideilty digital twin.\\

 \noindent Although NARGP has been validated in benchmark problems, its application in real-world scenarios has not been widely explored. To demonstrate the effectiveness of our framework, we validated it on a full-scale net cage at the \href{https://www.sintef.no/en/all-laboratories/ace}{SINTEF ACE} fish farm in Norway, a leading laboratory facility for developing and testing new aquaculture technologies. Data for this study were sourced from the FhSim software \cite{reite2014fhsim} and field sensor measurements. Leveraging both HF and LF data, our framework ensures that predictions closely align with real observations. This framework accurately predicts the net cage's displacement at specific locations and the mooring line loads under current excitations. Additionally, in our study, we explore different ML methods for developing surrogate models to predict the entire net cage deformation. We compare the Gaussian process (GP) with a prior principal component analysis (PCA) step and graph convolutional networks (GCNs) \cite{rasmussen2006gaussian, kipf2016semi}. While GCNs are widely used in life, physical, and material sciences, our research demonstrates their effectiveness in handling complex structural dynamics \cite{li2023application}.

\section*{Results}
Our goal is to develop a framework for a digital twin that enables real-time monitoring and visualization of aquaculture net cage dynamics under stochastic sea conditions, ensuring efficiency, safety, and cost-effectiveness while minimizing environmental impacts. Machine-learned surrogate modeling is fundamental in order to create the map between the marine conditions and the net cage dynamic responses. Accurate prediction of real-world behaviors relies on comprehensive, high-quality data, but acquiring such data is often challenging or cost-prohibitive. Building reliable models becomes complex when only a limited number of high-fidelity observations, such as direct sensor measurements, are available. This limitation often leads to the use of models that represent the real system based on more accessible data, such as those from simplified numerical models. However, a significant gap typically exists between these simulated outcomes and actual sensor readings. Additionally, it is crucial that this model can be trained and executed rapidly to function as an on-the-fly digital twin. Therefore, employing less complex methods is essential to achieve fast performance.\\

\noindent We propose a framework that utilizes data from various fidelity sources to learn the system dynamics, producing outputs that closely match real-world observations.
At the core of this framework is the NARGP method, as described by Perdikaris et al. (2017) \cite{perdikaris2017nonlinear}, which refines low-fidelity numerical simulation solutions to more closely match high-fidelity sensor data by learning the nonlinear correlations between these two data sources using data-efficient multifidelity information fusion algorithms and the simple GP methods. The NARGP method employs a sequential training process across each fidelity level. Detailed description of the NARGP is provided in the section Methods.
For our case study, which is detailed in the next section, the model is designed with two fidelity levels but is scalable to include additional layers from different data fidelities if necessary. The trained NARGP model is implemented in the digital twin following the workflow illustrated in Figure \ref{fig:DT-overview}. It receives real-time metocean data, and the GP model, trained on the low-fidelity numerical data, predicts the net cage dynamics. These low-fidelity predictions, along with the metocean data, are then input into the recursive GP surrogate model. This model has been trained using the GP posterior mean derived from the low-fidelity model and a few high-fidelity data, and it is able to provide predictions that accurately correspond to the actual net cage response.\\

\noindent The proposed multifidelity framework enables the use of simple, low-cost models that may have lower accuracy, and significantly enhances their precision by incorporating a small set of high-fidelity observations. By leveraging the nonlinear cross-correlations between low- and high-fidelity data through machine learning, this approach achieves substantial computational efficiency. This method allows us to tackle complex problems that would be infeasible to address using only high-fidelity data, which are often difficult to obtain due to cost and other limitations.

\subsection*{Case study for SINTEF ACE fish farm}
The multifidelity digital twin for real-time monitoring of net cage structural dynamics is built and validated for a fish farm at \href{https://www.sintef.no/en/all-laboratories/ace}{SINTEF ACE}, which is a full-scale laboratory facility designed to develop and test new aquaculture technologies. All relevant data used in this study correspond to this farm.

\subsubsection*{Data Collection}

\begin{figure}[ht]
\centering
\includegraphics[width=\linewidth]{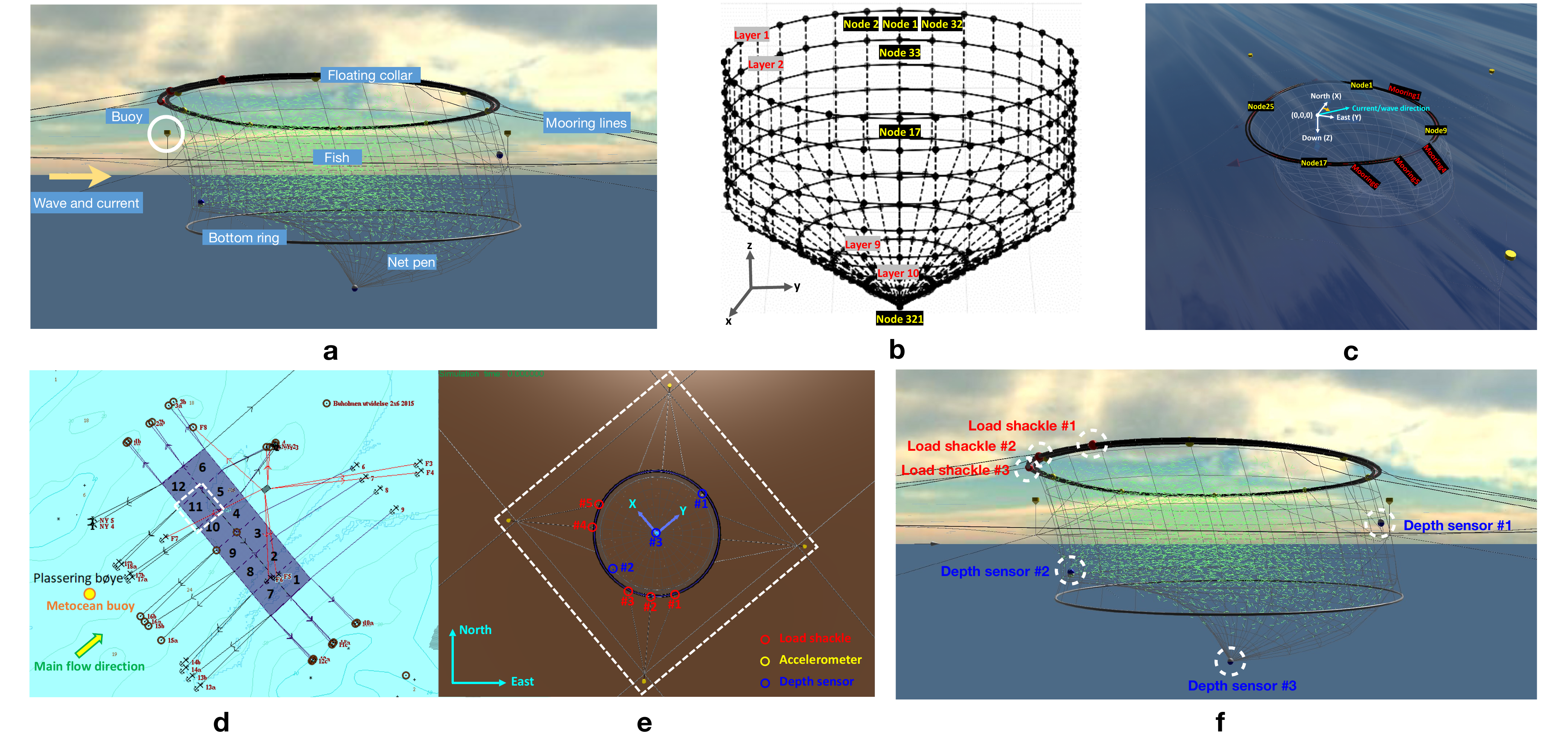}
\caption{\textbf{Net cage numerical representation and field sensor placement}. \textbf{(a)} Detailed view of net cage's critical components. \textbf{(b)} The net cage is discretized into 321 nodes for FhSim numerical simulations. It consists of 10 layers, with the first 32 nodes forming the first layer. Node 321 is at the bottom, connecting all nodes of the 10th layer. \textbf{c)} The net cage in the FhSim simulation environment, showing deformation under wave and current excitation. \textbf{(d)} Top view of the fish farm highlighting net cage 11, which accommodates the sensors. A buoy located 400 meters from the net cage measures waves and currents, with the main flow direction indicated. \textbf{(e)} Close-up of net cage 11 showing sensor locations: 5 load shackles measuring mooring line forces and 3 depth sensors measuring net displacement. \textbf{(f)} Side view of the net cage illustrating sensor locations along its depth.
}
\label{fig:FhSim-Sensors}
\end{figure}

\noindent\textbf{Low-fidelity data}. FhSim is a software platform and framework for mathematical modeling and numerical simulation, with a focus on marine applications, providing a time-domain representation of complex system dynamics. In this study, FhSim is used to model the dynamic response of a full-scale net cage to marine environment. Figure \ref{fig:FhSim-Sensors} (a) displays the numerical net cage in the FhSim environment, detailing its critical components. The numerical net cage has the same dimensions as the real cage, featuring a 50 m diameter, 18 m depth, and a 0.21 solidity ratio, attached by 12 bridle lines to the mooring frame. Although the actual net cage comprises thousands of twines, for numerical simplification, the net cage is discretized into 321 nodes, as shown in Figure \ref{fig:FhSim-Sensors} (b). In FhSim simulations, the displacements of each of the 321 nodes under wave and current excitation are calculated, fully capturing the net cage's deformation, as shown in Figure \ref{fig:FhSim-Sensors} (c).\\

\noindent For the purpose of this study, we assessed the net cage's dynamic behavior under 1,000 sea conditions covering a broad spectrum of wave and current conditions. The environmental parameters included current velocities ranging from 0 to 1 m/s, significant wave heights from 0 to 3 m, peak wave periods from 0 to 8.66 s, and wave and current directions from 0 to 360 degrees. Each simulation modeled the net cage interaction with the sea conditions over a 30-minute period, providing as output the $x$, $y$, and $z$ displacements for each of the 321 nodes and the loads on the 12 mooring lines. We focused on estimating the average value of the quantities of interest under each examined sea condition.\\

\noindent FhSim was previously used by Su et al. (2023) \cite{su2023towards} to simulate net cage dynamics under measured wave and current conditions in the fish farm. The direct solution of the FhSim model was found to underestimate net deformations, and this was partially attributed to the increased net solidity due to biofouling, as well as the possible alterations of the current from the measuring point (i.e., 400 m away). The authors\cite{su2023towards} developed and integrated an adaptive model that makes connections to the environmental condition inputs so that the calculated structural dynamics match observations. The FhSim's direct solution aims to predict the steady state of net cage's dynamics, rather than all the transient responses; therefore, it is viewed in our study as low-fidelity source of data. Apart from the solution accuracy, conducting FhSim simulations requires computational resources and time. Specifically, in our study, a 30-minute simulation time took approximately 10 minutes of CPU time (utilizing 56 cores, 2.2 GHz, and 128 GB RAM), with an additional 20 minutes required for processing and storing time-series data for all mooring lines and nodes. This makes it impractical for direct implementation in a digital twin for real-time monitoring, where immediate knowledge of the net cage response is essential.\\

\noindent\textbf{High-fidelity data}. 
Field sensor measurements capture the actual response of the system and are considered high-fidelity data. As shown in Figure \ref{fig:FhSim-Sensors} (d), a metocean buoy, positioned 400 meters away from the fish farm, was equipped with sensors to monitor the sea conditions, specifically measuring incoming waves and currents. Sensors were placed in one of the net cages at SINTEF fish farm, specifically in cage number 11 (Figure \ref{fig:FhSim-Sensors} (d)). The sensors included five load shackles (\#1-5 in Figure \ref{fig:FhSim-Sensors}(e)) installed between the collar and the mooring lines on the upstream side, where the prevailing water current originated, to measure mooring line loads. Depth sensors were installed at three different layers: 7 meters deep, at the boundary of the cylindrical part (15 meters), and at the bottom (31 meters) of the net pen (\#1–3 in Figures \ref{fig:FhSim-Sensors} (e) and (f)). These depth sensors measured the vertical displacement of the net cage at their respective positions, providing critical insights into the net cage deformation. The data were collected using a wireless sensor network and communication system over a monitoring period of slightly less than two months, from January 18 to March 5, 2020. This period captures conditions representative of a relatively harsh time of year. Data points are available at a high temporal resolution of 2-second intervals, providing comprehensive insights into the variations and dynamics of the measured quantities. Continuous measurements were recorded by the metocean buoy and the three depth sensors covering the entire period from January 18 to March 5, 2020. However, load shackle data were only available for a total of 36 hours—12 hours on January 23 and 24 hours on February 22, highlighting the risk of data loss that field sensors may experience. Detailed information about the field deployment and sensor measurements is available in the publication by Su et al. (2023) \cite{su2023towards}.

\subsubsection*{Digital twin workflow}

\begin{figure}[ht!]
\centering
\includegraphics[width=0.9\linewidth]{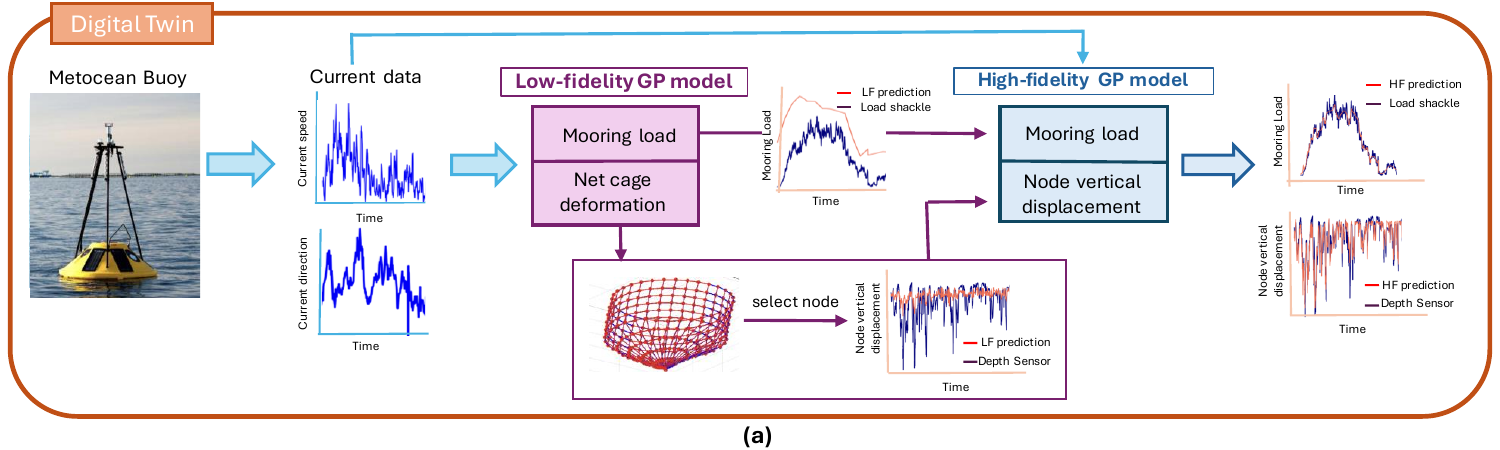}
\includegraphics[width=0.9\linewidth]{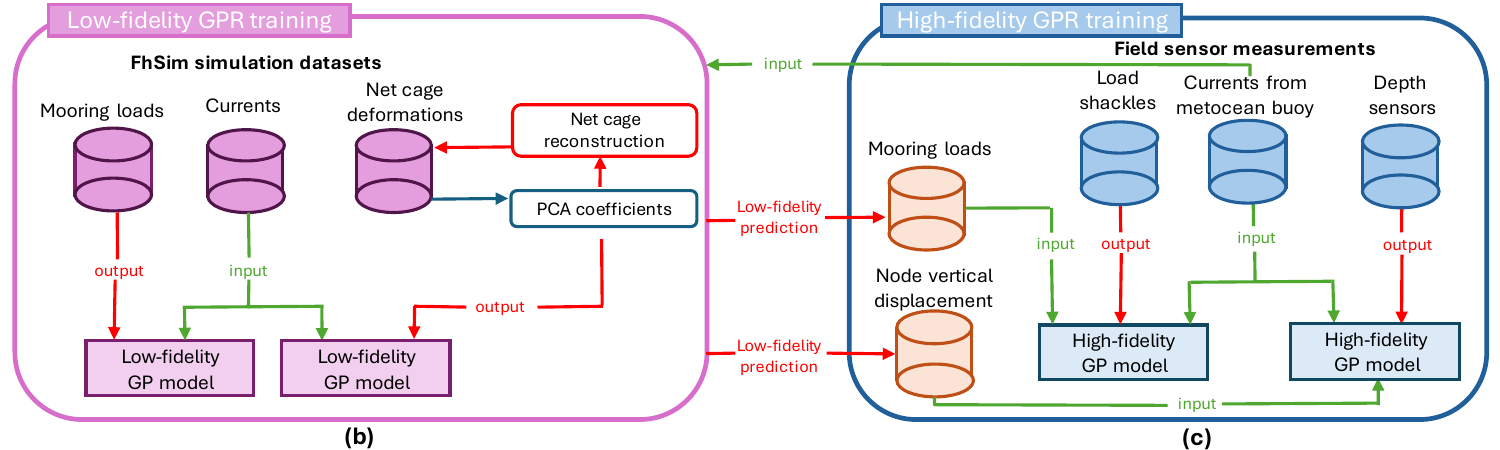}
\caption{\textbf{Multifidelity digital twin for SINTEF ACE fish farm.} \textbf{(a)} The digital twin's workflow is illustrated. 
\textbf{(b)} The training stage of the low-fidelity GP models using data from FhSim simulations is described. 
\textbf{(c)} The training stage of the high-fidelity GP models based on multifidelity data is illustrated. 
}
\label{fig:DT}
\end{figure}

\noindent Figure \ref{fig:DT} (a) illustrates the digital twin workflow developed for the SINTEF ACE fish farm. A metocean buoy equipped with sensors provides real-time measurements of current speed and direction which serve as inputs to the NARGP multifidelity surrogate model. Initially, the low-fidelity GP models use the current data to predict the loads on the mooring lines and the overall deformation of the net cage under these conditions. To estimate the net cage displacement at specific locations, the user defines these locations (or nodes) on the net cage and obtains their vertical displacement. The low-fidelity predictions of mooring line force and node vertical displacement, along with the measured current speed and direction, are then fed into high-fidelity GP models. These models refine the low-fidelity predictions, providing outputs that closely match the real structural dynamics measured by load shackles and depth sensors. This workflow is completed in a few seconds due to the algorithmic simplicity and low computational cost of the NARGP method. The proposed digital twin enables real-time monitoring of net cage structural dynamics, supports autonomous operations, and informs decision-making without relying on real-time sensors from the net cage, as is currently the case\cite{su2023towards}. This approach reduces the need for extensive sensor deployment, which is often expensive and unreliable due to issues like data loss. \\

\noindent Figure \ref{fig:DT} (b) shows the training stage of the low-fidelity GP surrogate models using data from FhSim simulations, described in the "Data Collection/Low-fidelity" section. Using the standard GP method\cite{rasmussen2006gaussian}, we build surrogate models to map current speed and direction to the mooring line load. To create the surrogate model that maps ocean currents to net cage deformation, dimension reduction using PCA is required as a preliminary step, as detailed in the "Methods/Net cage displacement data dimension reduction using PCA" section. This is necessary due to the extensive data describing net cage deformation ($x$, $y$, $z$ displacements for each of the 321 nodes). From the PCA dimension reduction step, we found that three PCA coefficients were sufficient and this allowed us to develop a standard GP model to map current characteristics to these PCA coefficients. Once the predicted coefficients are obtained, a reconstruction stage follows to determine the net cage deformation under the incoming measured current, as described in the "Methods/Machine learning functional relationships between current and PCA coefficients" section. Having the entire geometry of the deformed net cage allows us to extract the vertical displacement at the locations where the depth sensors are placed. The low-fidelity GP models developed at this stage can fully substitute FhSim simulations, with their accuracy well-documented by Katsidoniotaki et al. (2024) \cite{katsidoniotaki2024}. This ensures that if the load shackles and depth sensors change position or more sensors are added in the future, retraining the low-fidelity GP models will not be necessary.\\

\noindent Figure \ref{fig:DT} (c) outlines the steps for training the high-fidelity GP models following the NARGP method, as described in the "Methods/NARGP" section. The high-fidelity training data consist of field sensor measurements, including current measurements from the metocean buoy, mooring loads from load shackles, and net cage vertical displacements from depth sensors, as detailed in the "Data Collection/High-Fidelity" section. The current measurements are input into the low-fidelity GP models, and the predicted GP posterior mean, combined with the current measurements, serve as inputs for the high-fidelity models. The high-fidelity GP models are trained on a limited number of datasets obtained from the period January 18 to March 5, 2020, to predict quantities of interest that match the load shackle and depth sensor measurements. Specifically, while the depth sensor measurements cover the entire period, only a small number of datasets corresponding to a few minutes are used to train the model. For the load shackle measurements, available only for 36 hours, datasets corresponding to a few seconds are used for training. This approach demonstrates how, with limited high-fidelity datasets, we can develop a multifidelity model that provides predictions representative of real observations.

\subsubsection*{Predicting net cage dynamics}

\begin{figure}[ht!]
\centering
\includegraphics[width=0.8\linewidth]{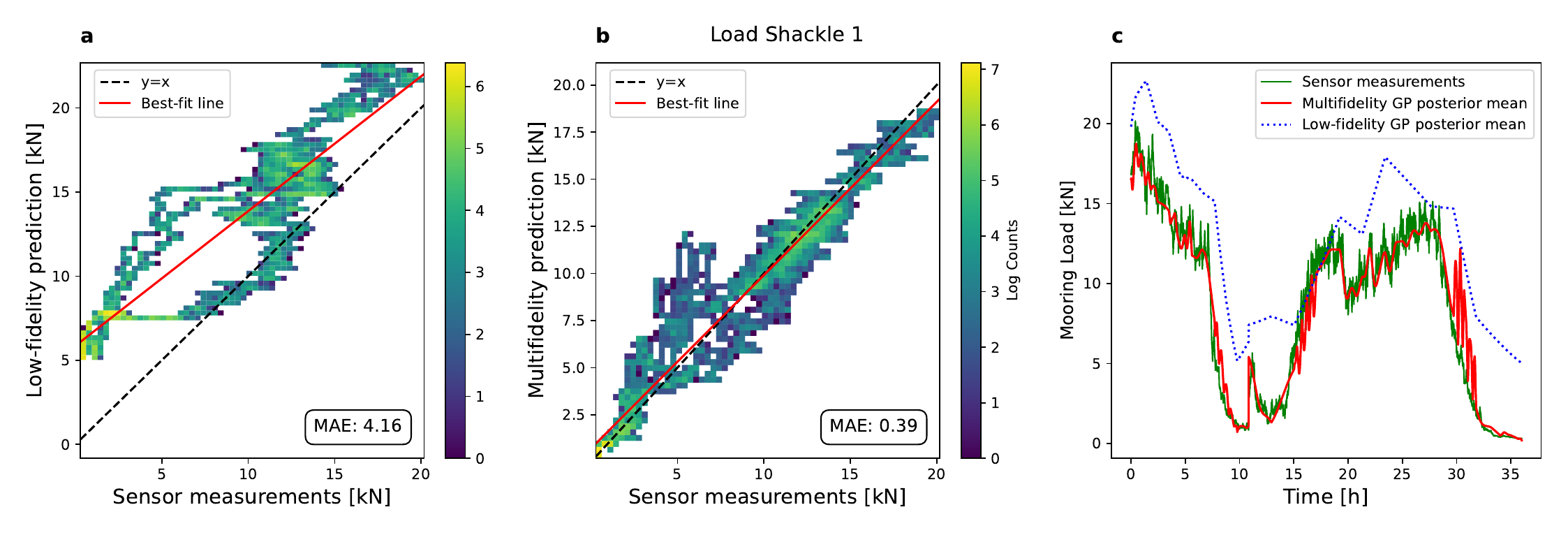}
\includegraphics[width=0.8\linewidth]{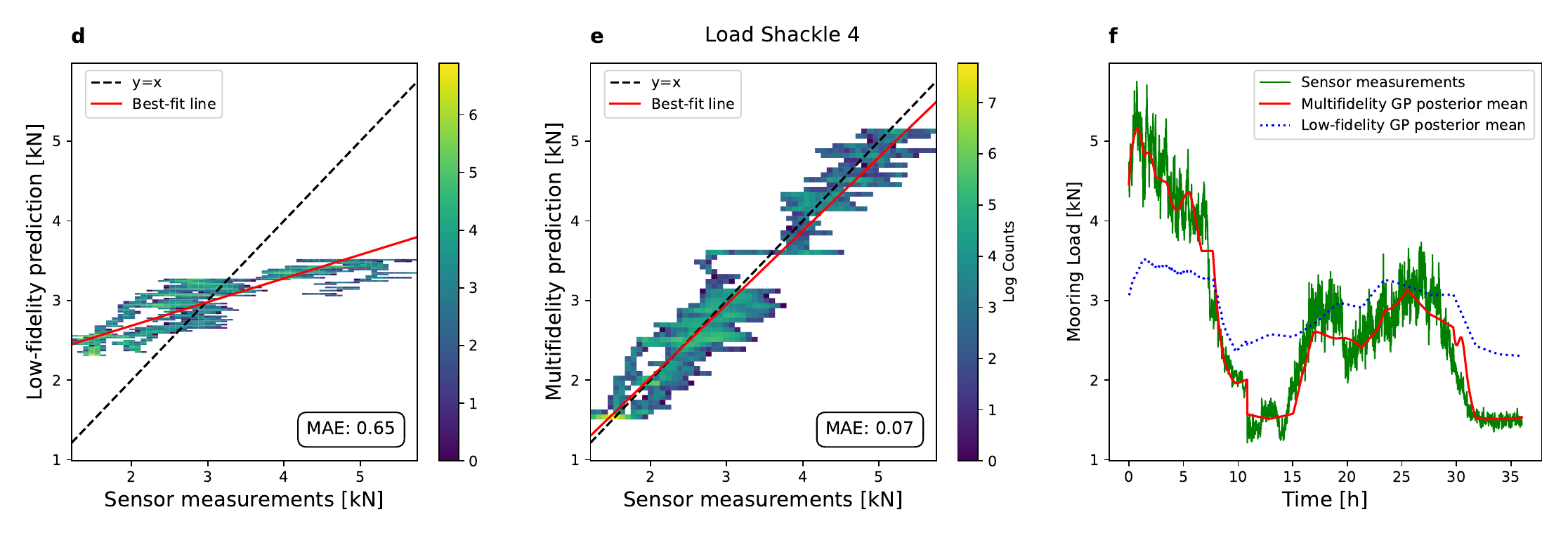}
\includegraphics[width=0.8\linewidth]{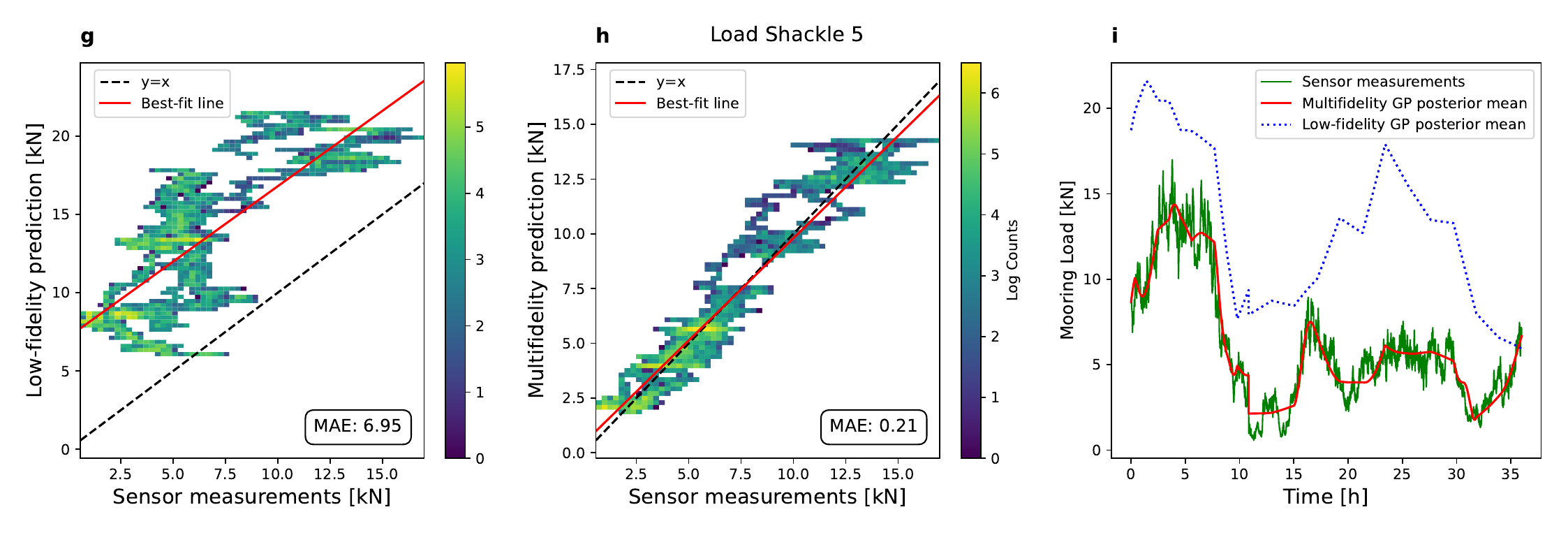}
\caption{
\textbf{Multifidelity framework predictions for mooring line loads.} Scatter plots \textbf{(a), (d), (g)} show the low-fidelity GP posterior mean against load shackle measurements. Color scale indicates log count density of data points, with lighter colors representing higher concentrations. The dashed diagonal line signifies perfect prediction accuracy; the red line is the best-fit trend. 
Scatter plots \textbf{(b), (e), (h)} show the multifidelity GP model predictions against sensor measurements. 
Plots \textbf{(c), (f), (i)} display predictions over a 36-hour period, excluding the data used for training. Low-fidelity GP posterior mean deviates significantly from actual measurements. Multifidelity GP posterior aligns closely with real observations, capturing the overall trend despite some oscillatory behavior deviations. 
}
\label{fig:LS-results}
\end{figure}

\noindent\textbf{Mooring line loads}. Figure \ref{fig:LS-results} showcases the proposed multifidelity framework's proficiency in predicting mooring line loads under current conditions, as measured by the metocean buoy, and compares these predictions with field measurements obtained from load shackles. The figure presents predictions for load shackles \#1 (top row), \#2 (middle row), and \#5 (bottom row), with similar patterns observed for other mooring lines. Validation was conducted using previously unseen data from the entire 36-hour load shackle measurement period, excluding the training datasets. The 2D histograms in Figure \ref{fig:LS-results} are scatter plots with a color scale depicting the distribution and density of prediction accuracy from the low-fidelity and multifidelity GP models. The color scale indicates the log count density of data points, with lighter colors representing areas with a higher concentration of data points. Proximity to the dashed diagonal line ($y=x$) signifies prediction accuracy, while the red line represents the best-fit trend. The scatter plots for the low-fidelity GP model (Figure \ref{fig:LS-results} (a), (d), (g)) reveal a broader spread and a significant deviation of the red best-fit line from the dashed diagonal line, indicating less accurate predictions. Conversely, the multifidelity GP model plots (Figure \ref{fig:LS-results} (b), (e), (h)) exhibit a higher concentration of lighter colors near the dashed line, with the red best-fit line aligning more closely with the diagonal, indicating improved prediction accuracy. The color gradient helps identify trends and outliers; consistent, lighter colors along the dashed line suggest the model captures the overall trend well, whereas significant deviations indicate less accurate predictions. The Mean Absolute Error (MAE) quantifies prediction accuracy, calculated as the average of the absolute differences between the predicted and actual load measurements. The multifidelity model demonstrates a significant improvement in MAE values compared to the low-fidelity model. Figure \ref{fig:LS-results} (c), (f), (i) provides a temporal view, displaying predictions over time based on the validation data from the 36-hour measurement period. The low-fidelity GP posterior mean deviates notably from the actual measurements, whereas the multifidelity GP posterior aligns closely with the real observations. The temporal plots reveal an oscillatory behavior in the load shackle measurements not fully captured by the multifidelity model, explaining some deviations from the dashed line in the middle column's scatter plot. Nonetheless, the multifidelity model accurately follows the overall trend of the observations. \\ 

\begin{figure}[ht!]
\centering
\includegraphics[width=0.8\linewidth]{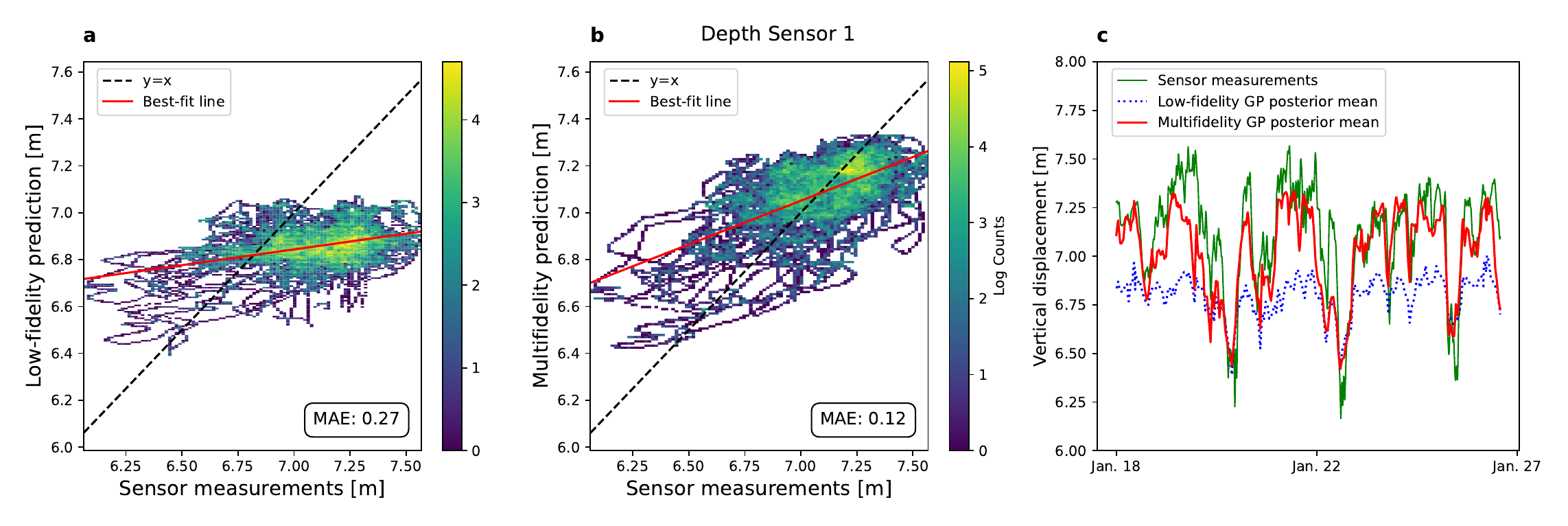}
\includegraphics[width=0.8\linewidth]{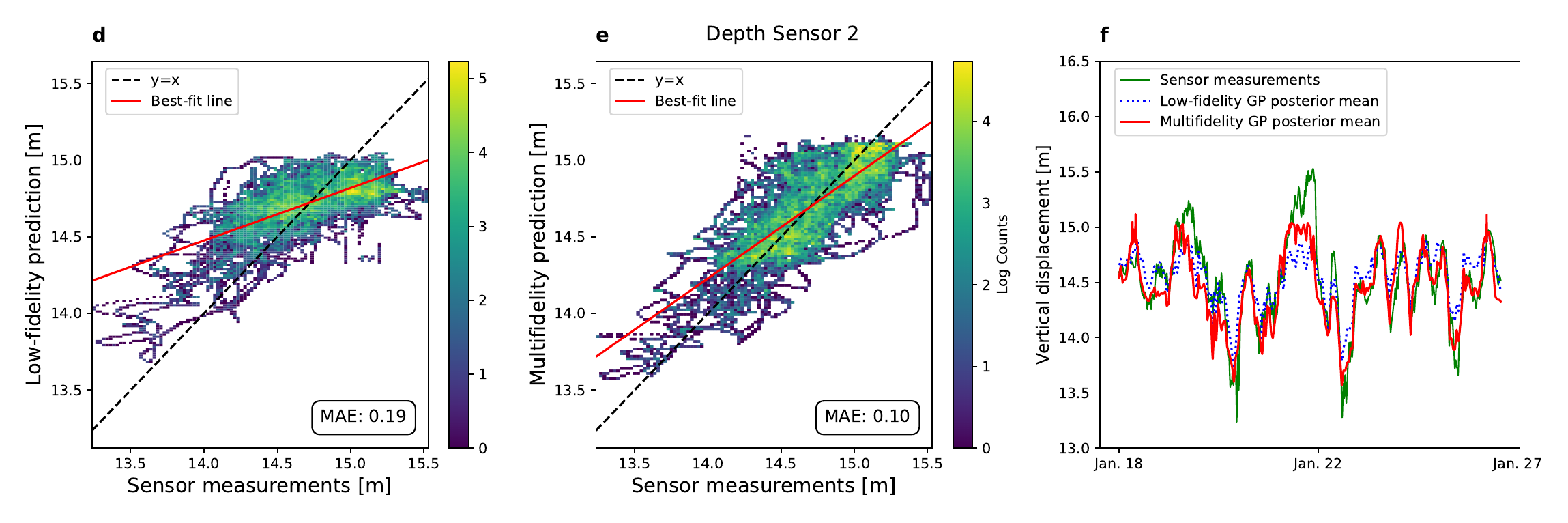}
\includegraphics[width=0.8\linewidth]{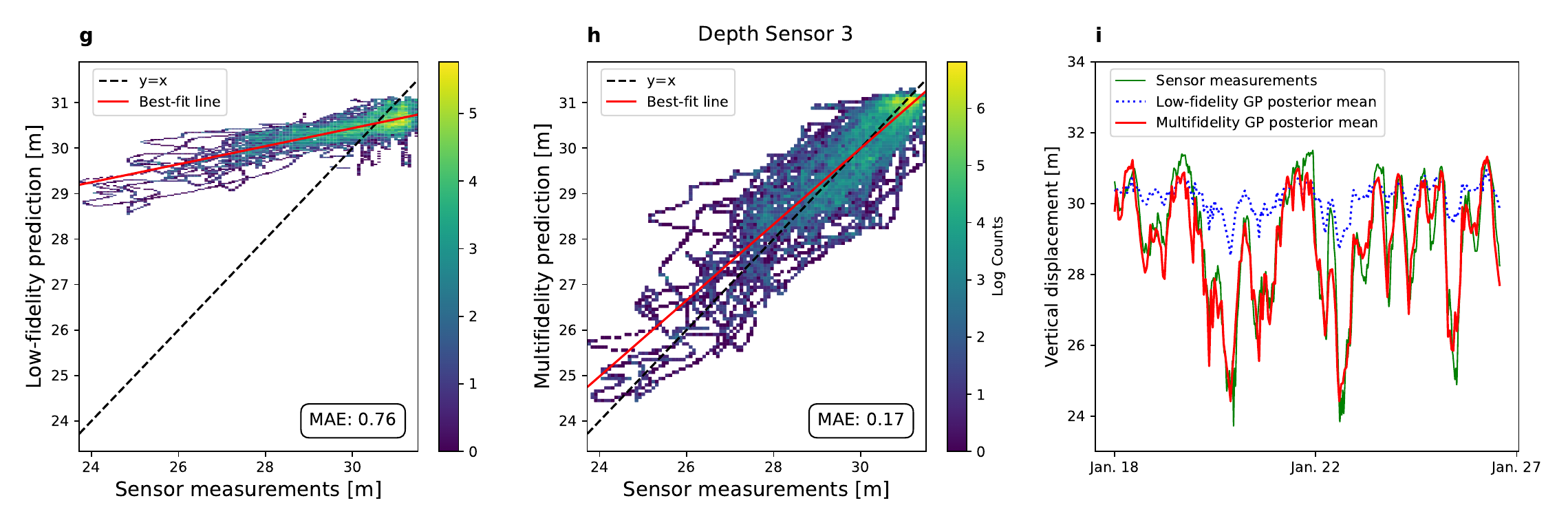}
\caption{
\textbf{Multifidelity framework predictions for net cage displacement.} Scatter plots \textbf{(a), (d), (g)} contrast the predictions from the low-fidelity GP model with the depth sensor measurements. The broad dispersion of data points and the notable deviation of the red best-fit line from the dashed diagonal line underscore the model's limited accuracy. 
Scatter plots \textbf{(b), (e), (h)} for the multifidelity GP model. Data points cluster more closely around the dashed line, and the red best-fit line aligns more precisely with the diagonal, indicating enhanced predictive accuracy. 
The MAE values annotated on these plots quantitatively highlight the multifidelity model's superiority over the low-fidelity model.  Plots \textbf{(c), (f), (i)} show the temporal comparison of predicted and actual net cage displacements over a specific period (January 18 to January 27, 2020). The low-fidelity GP model (blue line) exhibits substantial deviations from the actual measurements (green line), particularly for depth sensors \#1 and \#3. The multifidelity GP model (red line) significantly mitigates these discrepancies, closely mirroring the actual measurements and capturing the underlying trend with greater fidelity.
}
\label{fig:DS-results}
\end{figure}

\noindent\textbf{Vertical displacement at specific points within the net cage.} 
 Figure \ref{fig:DS-results} illustrates the predictive performance of the multifidelity framework in estimating net cage displacement at three distinct locations, as measured by depth sensors \#1 (top row), \#2 (middle row), and \#3 (bottom row). The validation dataset spans from January 18 to March 5, 2020, excluding the training periods.  Figure \ref{fig:DS-results} (a), (d), (g) presents scatter plots contrasting the predictions of the low-fidelity GP model with actual sensor measurements. The broad dispersion of data points and the notable deviation of the red best-fit line from the dashed diagonal line underscore the model's limited accuracy. In contrast, Figure \ref{fig:DS-results} (b), (e), (h) displays scatter plots for the multifidelity GP model. Here, data points cluster more closely around the dashed line, and the red best-fit line aligns more precisely with the diagonal, indicating enhanced predictive accuracy. Despite the inherent complexities and dynamic motion of the flexible net cage, the multifidelity model's predictions, though still dispersed, show improved accuracy, as evidenced by the color scale: darker points represent outliers, while lighter points closer to the dashed line signify higher accuracy. The MAE values annotated on these plots quantitatively highlight the multifidelity model's superiority over the low-fidelity model. Figure \ref{fig:DS-results} (c), (f), (i) provides a temporal comparison of predicted and actual net cage displacements over a specific period (January 18 to January 27, 2020), juxtaposing the low-fidelity GP and multifidelity GP posterior means against the actual sensor data. The low-fidelity GP model (blue line) exhibits substantial deviations from the actual measurements (green line), particularly for depth sensors \#1 and \#3. Conversely, the multifidelity GP model (red line) significantly mitigates these discrepancies, closely mirroring the actual measurements and capturing the underlying trend with greater fidelity.  

\subsubsection*{Flexible net cage deformation via surrogate modeling}

Fish farm net cages are flexible structures that tend to follow rather than resist water motions, therefore they present significant deformations. These deformations can impact fish survival and the structural integrity of the cages, with any failure potentially leading to fish escapes. This underscores the importance of monitoring net cage deformations to maintain a comprehensive understanding of fish farm conditions.
Due to their inherent flexibility, monitoring the entire deformation of net cages using available sensors is challenging. Depth sensors can be mounted strategically on the net structure thereby obtaining a 3D position of these points in the net, which in turn are used to extrapolate the full net structure using mathematical models\cite{fore2024digital}. A similar approach was implemented by Su et al.\cite{su2023towards} for the development of a digital twin for real-time monitoring of aquaculture net cage systems, where depth sensor data were assimilated into FhSim simulations to represent the actual net cage system. Input properties for the simulations were adapted to ensure the numerical simulation outputs fit the sensor-measured values. However, this approach depends on sensors transmitting position information, with sensors being expensive and prone to data transmission loss. Additionally, it requires substantial computational time to process the data and calculate the correct net cage deformation. These reasons make this approach less suitable for real-time monitoring and immediate assessment of the system's state.\\

\begin{figure}[ht]
\centering
\includegraphics[width=0.225\linewidth]{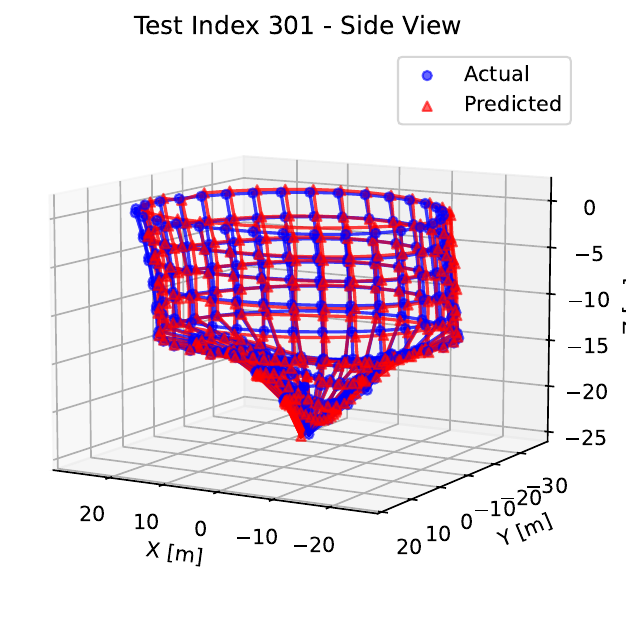}
\includegraphics[width=0.216\linewidth]{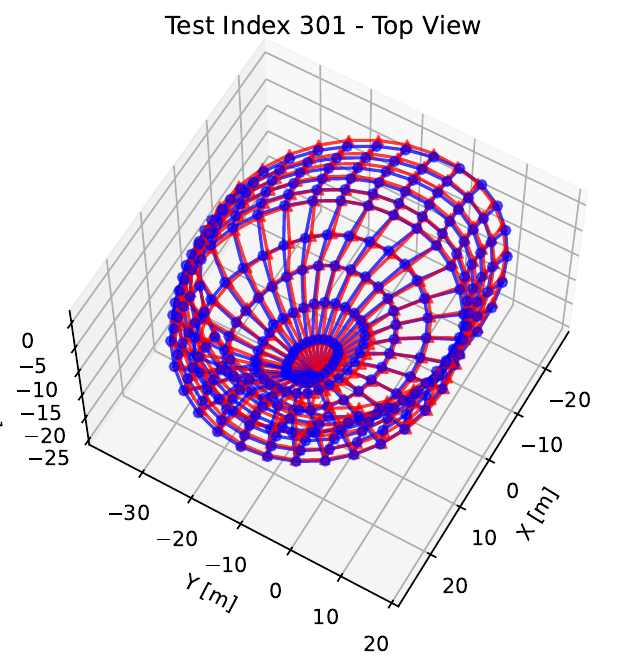}
\includegraphics[width=0.225\linewidth]{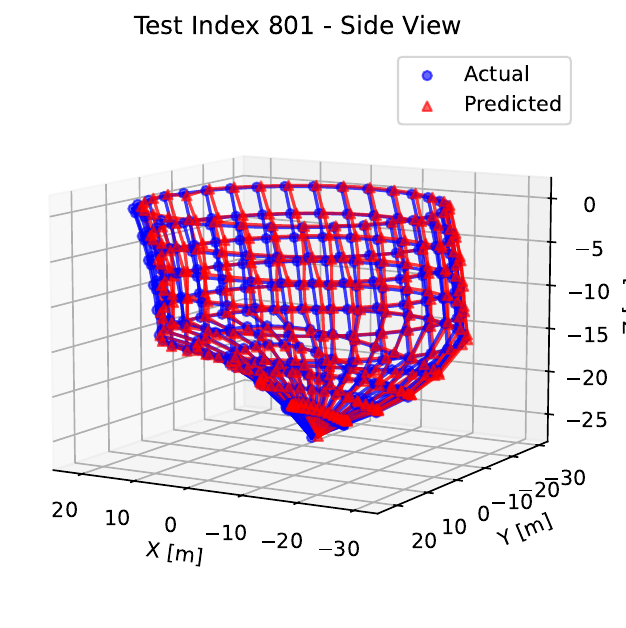}
\vspace{0.4cm}
\includegraphics[width=0.216\linewidth]{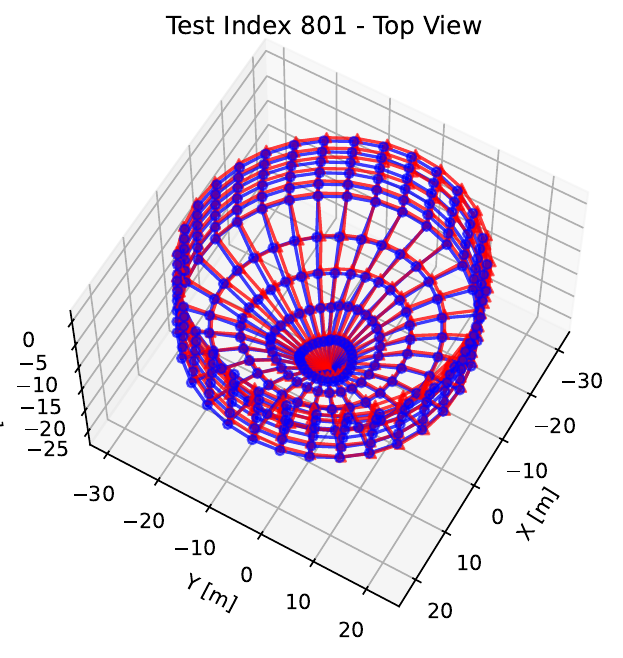}
\includegraphics[width=0.9\linewidth]{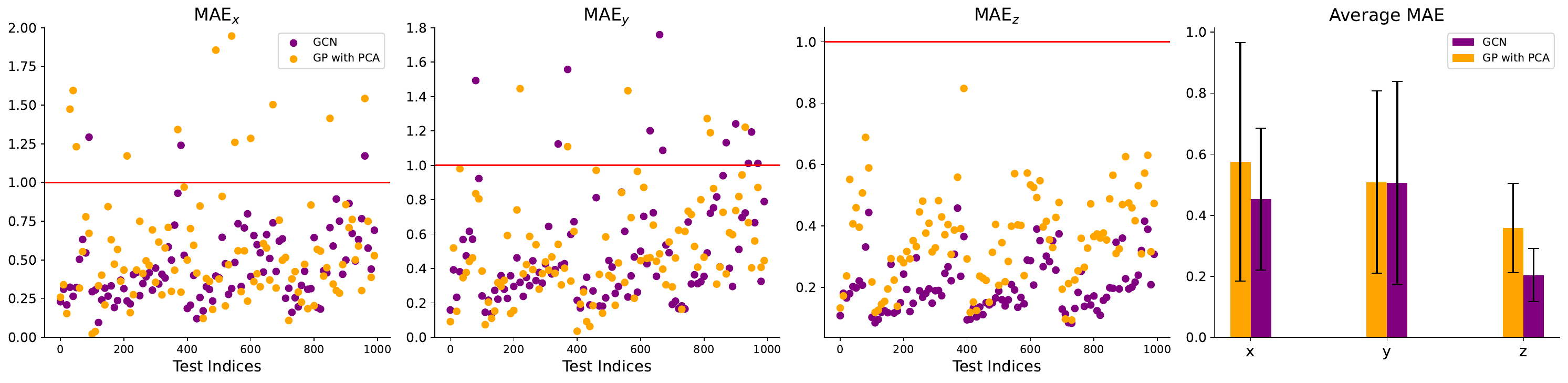}
\caption{\textbf{Machine-learned surrogate models for predicting net cage deformation.} \textbf{(a)} Comparison of net cage deformation predictions under harsh (301) and mild (801) current excitations by Graph Convolutional Networks (GCNs) with FhSim simulation solutions. \textbf{(b)} Evaluation of net cage deformation prediction accuracy using MAE metric, comparing predictions of each node displacement in the $x$, $y$, $z$ directions with the FhSim solution. Two surrogate modeling methods are compared: (i) GP surrogate with prior PCA, and (ii) GCNs. The red line at MAE = 1 serves as a reference point, indicating the threshold beyond which prediction error significantly impacts operations such as UUV control. The surrogates are evaluated across 100 test cases. The final plot summarizes the MAE, showing the average magnitude of errors in predictions for each axis, with error bars representing the variability in the error.
}
\label{fig:MAE}
\end{figure}

\noindent Our study previously introduced the multifidelity digital twin, which accurately reflects the real net cage response to the dynamic marine environment. As a result, this digital twin eliminates the reliance on real-time data measured by sensors placed in the cage. Additionally, as illustrated in Figure \ref{fig:DT} (b), we developed surrogate models, using the standard GP method with a prior PCA, that map the net cage deformation with the currents. The accuracy of these models, previously presented by Katsidoniotaki et al.\cite{katsidoniotaki2024}, suggests that they can replace FhSim simulations and significantly reduce computational costs in digital twin applications.
However, surrogate models inherently introduce some degree of prediction error, depending on the method used to build them. This error can impact autonomous operations or decision-making where good accuracy is required. In this study, we compare the GP-PCA surrogate model with a surrogate model built using GCNs to predict net cage deformation topology under varying sea conditions. 
We chose the GCN since the net cage resembles graph-structured data. The cage is discretized into 321 nodes, which are connected in the manner shown in Figure \ref{fig:FhSim-Sensors} (b). The GCN receives node features and the topology of their connections and can predict the displacement of each node under dynamic marine conditions. In the section "Methods/Graph Convolutional Networks for Net Cage Deformation," further details about the GCN model are provided. Figure \ref{fig:MAE} compares the predicted net cage deformation from the GCN model (red) with the deformation estimated by FhSim simulations (blue) under both mild (test index 801) and harsher sea conditions (test index 31).\\

\noindent To evaluate the prediction error of each method, we consider the example of autonomous navigation of Unmanned Underwater Vehicles (UUVs), which require real-time information about the surrounding environment where they navigate. During autonomous operations, a UUV such as a Remotely Operated Vehicle (ROV) should maintain a distance of at least 1 meter from the net cage to ensure collision-free motion. Therefore, it is imperative to keep the prediction error regarding net cage deformation within acceptable limits for safe operations.  
The MAE metric was employed to quantify the accuracy of net cage deformation predictions across three spatial directions over a range of test indices. As depicted in Figure \ref{fig:MAE}, the majority of data points for both methods lie below the red line at MAE = 1 meter, indicating a satisfactory level of accuracy for most test indices. Notably, higher MAE values are observed in test cases involving higher current velocities, where the net cage exhibits larger deformations, highlighting the challenges in modeling more extreme conditions. As summarized in the final subplot in Figure \ref{fig:MAE}, the GP-PCA model demonstrates higher MAE values, suggesting lower prediction accuracy compared to the GCN model. The wider spread of the GP-PCA error bars in the $x$ and $z$ dimensions notably suggests greater variability in its predictions. This greater error in the GP-PCA model may be attributed to dual sources of error: the PCA dimension reduction, which entails some loss of information, and the inherent errors in the GP model itself.

\section*{Discussion}

In this study, we present a framework that leverages multifidelity surrogate modeling capabilities, designed to be integrated into a digital twin for real-time monitoring of structural dynamics. Unlike traditional digital twins that rely on physics-based simulations—requiring significant computational resources and thus being suboptimal for real-time monitoring—our approach offers several key advantages. Our proposed framework requires less computational cost during both the training and utilization stages, allowing it to operate on the fly and provide rapid predictions. Furthermore, our framework effectively addresses a common challenge in real-world applications: the limited availability and access to real-world data for developing realistic predictive surrogate models. By seamlessly integrating low-fidelity numerical data with high-fidelity field sensor measurements, our model bridges the gap between simulation outcomes and real-world observations. This integration enhances the accuracy and reliability of predictions, making the digital twin highly suitable for real-time monitoring applications, even in scenarios where real-world data is scarce.\\

\noindent The NARGP method, central to our framework, belongs to a class of multifidelity information fusion algorithms. An important feature of the NARGP method is the nonlinear autoregressive scheme that enables robust correlation among different fidelity data sources, addressing the challenges posed by nonlinear relationships. While NARGP has been validated against benchmark problems, it has not yet been validated for a real application. The NARGP method performs well in our application, providing good prediction accuracy by significantly correcting the low-fidelity solution to match the real solution quickly and at a low computational cost. \\ 

\noindent The framework successfully validated for the test case of SINTEF ACE fish farm, using low-fidelity numerical simulation data and high-fidelity field sensor measurements. For the net cage's vertical displacement at specific locations, the low-fidelity GP prediction deviates significantly from the depth sensor measurements. This discrepancy is likely due to the pronounced nonlinear and complex phenomena associated with the hydrodynamic effects due to the flexible net cage's interaction with the wave and currents, which cannot be fully captured by numerical simulations. However, the NARGP method effectively corrects these deviations, resulting in a final prediction that closely aligns with the depth sensors. Our framework was developed and validated using high-fidelity data from field sensor measurements obtained over a period slightly less than two months, specifically from January 18 to March 5, 2020. Although this period covered harsh sea conditions, ideally, measurements should span the entire year to account for seasonal variations. In the future, if more field sensor data become available, the high-fidelity surrogate model can be retrained using new samples that cover chosen from the total available measurements, further improving the framework's accuracy and robustness.\\

\noindent Accurate predictions of mooring loads and net cage deformation are crucial for evaluating structural integrity, informing decisions on predictive maintenance, and preventing structural damage that could lead to fish mortality or escape. This digital twin also enables autonomous operations in fish farms, such as inspecting nets and mooring lines for damages, irregularities, biofouling conditions, and net cleaning\cite{kelasidi2023robotics}. These operations are currently performed using UUVs such as ROVs. For the autonomous navigation of these systems, it is essential to plan and control the motion of the ROV's motion to ensure collision-free movement in highly dynamic environments\cite{kelasidi2023robotics, kelasidi2022autonomous}. Since the ROV operates inside the net cage, real-time monitoring of net cage deformation is essential to avoid collisions.\\

\noindent Additionally, we offer a solution for the challenging task of predictive modeling of net cage deformation due to the structure's flexibility. Knowing the exact net cage deformation is critical for fish welfare, net cage structural integrity, and overall fish farm operations. Our study demonstrates the capability of surrogate modeling to predict the deformation of the entire net cage under diverse marine conditions. We showcase the effectiveness of GCNs in assimilating the graph-structured net cage data, achieving better prediction accuracy compared to the GP method with PCA dimension reduction. This study is one of the few to utilize GCNs for structural deformation problems, highlighting the potential for GCNs to be expanded beyond their current applications in fields such as drug discovery, biology, transportation, and computer vision.\\

\noindent In conclusion, in this study we introduce a solution aided by machine learning and multifidelity data assimilation for developing digital twins for the remote fish farm management, aiming to advance aquaculture industry with new practices. While our focus has been on predicting net cage deformation and mooring line loads, the multifidelity framework is generalizable for learning any stochastic nonlinear behavior of other quantities of interest in structural problems and other aquaculture matters, such as predicting fish behavior and environmental conditions in fish farms. The proposed digital twin solution improves efficiency and safety, prevents damages, reduces costs, and mitigates adverse environmental impacts. These factors are essential for the industry's expansion and the push towards offshore waters to meet rising food demands and secure food supply amid climate change.

\section*{Methods}

\subsection*{Nonlinear autoregressive GP regression (NARGP)}
The nonlinear autoregressive GP regression denoted by NARGP \cite{perdikaris2017nonlinear} is a class of multifidelity nonlinear information fusion algorithms that enables accurate inference of quantities of interest by synergistically combining realizations of low-fidelity models with a small set of high-fidelity observations, capable of learning complex nonlinear and space-dependent cross-correlations between models of variable fidelity. Our research adopts the NARGP method in response to the observed nonlinear dynamics between our low and high-fidelity datasets.\\

\noindent We have $s$ levels of information sources (fidelities) producing outputs $y_t(\mathbf{x_t})$, at locations $\mathbf{x_t} \in  D_t \subseteq\mathbb{R}^d$, with $ t = 1, \ldots, s$. In our application we have two fidelity levels, $s=2$, and two input parameters, $d=2$, i.e. current velocity and direction. We can organize the available datasets by increasing fidelity as $D_t = \{\mathbf{x_t}, \mathbf{y_t}\}$, while the datasets have nested structure, i.e. $D_2 \subseteq D_1$. This assumption implies that the training inputs of the higher fidelity level need to be a subset of the training inputs of the lower fidelity level.
In NARGP \cite{perdikaris2017nonlinear} method, the nonlinear autoregressive scheme reads as:
\begin{equation}
\label{eq:NARGP}
f_t(\mathbf{x}) = g_t(\mathbf{x}, f_{*_{t-1}}(\mathbf{x}))
\end{equation}
where $f_t$ is the GP modeling the data at fidelity level $t$, the $f_{*_{t-1}}$ is the GP posterior from the previous inference level $t-1$, and the function $g_t$ is modeled as a Gaussian prior, 
$g_t \sim \mathcal{GP (\mathbf{f_t}|\mathbf{0},\mathbf{k_{t_g}})}$, 
with a covariance kernel that composes as
\begin{equation}
\label{eq:kernel}
k_{t_g} = k_{t_\rho}(\mathbf{x}, \mathbf{x'};\theta_{t_\rho}) \cdot k_{t_f}(f_{*_{t-1}}(\mathbf{x}), f_{*_{t-1}}(\mathbf{x'}) ;\theta_{t_f}) + k_{t_\delta}(\mathbf{x}, \mathbf{x'};\theta_{t_\delta})
\end{equation}
The structure of the kernel $k_{t_g}$ reveals the effect of the deep representation encoded in Eq. \ref{eq:NARGP}. 
The $g_t(\mathbf{x}, f_{*_{t-1}}(\mathbf{x}))$ projects the lower fidelity posterior $f_{*_{t-1}}$ onto a $(d+1)$-dimensional latent manifold, that jointly relates the input space and the outputs of the lower fidelity level to the output of the higher fidelity model, from which we can infer a smooth mapping that recovers the high-fidelity response $f_t$. This allows to capture general nonlinear, non-functional and space-dependent cross-correlations between low- and high-fidelity data. In Eq. \ref{eq:kernel}, $k_{t_{\rho}}$, $k_{t_{f}}$, and $k_{t_{\delta}}$ are valid covariance functions and ${\boldsymbol{\theta_{t_\rho}}, \boldsymbol{\theta_{t_f}}, \boldsymbol{\theta_{t_\delta}}}$ denote their hyperparamters, which can be easily learnt from the data $\{{\mathbf{x_t}, \mathbf{y_t}}\}$ via the maximum likelihood estimation procedure followed by the standard GP \cite{rasmussen2006gaussian} using the kernel $k_{t_g}$. This approach requires the estimation of ($2d+3$) hyperparameters assuming that all kernels account for directional anisotropy in each dimension using automatic relevance determination (ARD) weights. The kernel functions are chosen to have the squared exponential form with ARD weights, i.e. 
\begin{equation}
k_t(x, x'; \theta_t) = \sigma_t^{2} \text{ exp } \left( -\frac{1}{2} \sum_{i=1}^d w_{i,t}(x_i - x_i')^{2} \right)
\end{equation}
where $\sigma_{t}^{2}$ is a variance parameters and $(w_{i,t})_{i=1}^{d}$ are the ARD weights corresponding to fidelity level $t$. These weights allow for a continues 'blend' of the contributions of each individual dimension in $\mathbf{x_t}$ as well as the posterior predictions of the previous fidelity level $f_{*_{t-1}}$, and they are learnt directly from the data when inferring $f_t$.\\

\noindent The first level of the proposed recursive scheme corresponds to a standard GP regression problem trained on the lowest fidelity data $\{\mathbf{x_1}, \mathbf{y_1}\}$, and therefore, the predictive posterior distribution is defined by standard's Gaussian mean and covariance using the kernel function $k_1(\mathbf{x_1}, \mathbf{x'_1};\theta_1)$. Figure \ref{fig:DT} provides a summary of the NARGP workflow as implemented in this study. 

\subsection*{Net cage displacement data dimension reduction using PCA}
PCA is a statistical technique used in data analysis for dimensionality reduction. It identifies the directions (or principal components) that maximize the variance in a dataset, thereby preserving essential information with fewer variables. In our study, we apply PCA to the dataset obtained from FhSim numerical simulations, which provide the $x_i$, $y_i$, $z_i$ displacements of the nodes into which the net cage is discretized, where $i=1, \ldots, 321$. We have this data for each of the $s=1000$ environmental conditions examined. Specifically, for each environmental condition, FhSim simulates the net cage interaction with waves and currents over a 30-minute period, outputting the $x_i(t)$, $y_i(t)$, $z_i(t)$ node displacements. The 30-min simulation time for all the $s=1000$ environmental conditions corresponds to a total of $t_n$ time steps. All the available data from these simulations are integrated into the matrix $\mathbf{M}$.
\[
   \mathbf{M} = 
   \begin{bmatrix}
     x_1(t_1) &  \dots & x_1(t_n) \\
     y_1(t_1) &  \dots & y_1(t_n) \\
     z_1(t_1) &  \dots & z_1(t_n) \\
      & \dots  \\
     x_{321}(t_1)  & \dots & x_{321}(t_n) \\
     y_{321}(t_1)  & \dots & y_{321}(t_n) \\
     z_{321}(t_1)  & \dots & z_{321}(t_n) \\
   \end{bmatrix}
\]
Using this data matrix, we perform eigenanalysis to calculate the eigenvectors, $\boldsymbol{\phi}$, and eigenvalues,$\lambda$, respectively.  
\begin{equation}
\label{eq:eigen}
\mathbf{M}\mathbf{M}^T \boldsymbol{\phi} = \lambda\mathbf{\phi}
\end{equation}
The eigenvalues and their corresponding eigenvectors are sorted in descending order of the eigenvalues. The eigenvector with the highest eigenvalue is the first principal component of the dataset and accounts for the highest variance, with each subsequent component capturing progressively less variance than its predecessor. We determine the number of principal components to retain based on the cumulative explained variance ratio, which translates into how many components $k$ we should keep out of the $c$ total components to better describe the data. This sum yields the total proportion of the dataset's variance that is accounted for by the first $k$ principal components. 
\begin{equation}
\label{eq:CVE}
\text{Cumulative Variance Explained} = \sum_{p=1}^{k} \frac{\lambda_p}{\sum_{j=1}^{c} \lambda_j}             
\end{equation}
A common approach is to choose the smallest possible number of principal components, $k$, that explain a substantial portion of the variance. However, the choice of $k$ is based on a predetermined threshold which reflects the desired level of variance explanation: $\text{Cumulative Variance Explained} \geq \text{Threshold}$. In our case, setting the threshold at 93\% results in the retention of $k$ = 3 principal components. \\

\noindent Next, we project the original data, $\mathbf{M}$, onto the selected principal components to transform them into the new feature subspace, by multiplying the original matrix by the first $k$ selected eigenvectors, to get the \(\textbf{B}\) matrix of the transformed data which can be referred as PCA coefficients. 
\begin{equation}
\mathbf{B} =   \mathbf{M} \cdot \phi          
\end{equation}
The resulting $\mathbf{B}$ matrix consists of 3 rows, each corresponding to one of the principal components, and $t_n$ columns each one corresponds to a time step state of the net cage deformation. Since we are interested in estimating the average shape of the net cage deformation for each of the $s= 1000$ scenarios, we estimate the mean principle component for each scenario, resulting in a matrix \(\mathbf{\bar{B}}\), which can be referred as mean PCA coefficients. This matrix has $k=3$ rows and $s =1000$ columns.
\[
   \mathbf{\bar{B}} = 
   \begin{bmatrix}
     \bar{b}_{1,1} &  \dots & \bar{b}_{1,s}  \\
     \bar{b}_{2,1} &  \dots & \bar{b}_{2,s}  \\
     \bar{b}_{3, 1} & \dots & \bar{b}_{3,s}  \\
   \end{bmatrix}
\]
To reconstruct the average net cage deformation for each of the total $s=1000$ environmental conditions, we multiple the eigenvector $\phi$ with the principal components. 
\begin{equation}
\label{eq:rec}
\mathbf{\bar{M}} = \sum_{h=1}^{s}\sum_{j=1}^k \phi_j \cdot \bar{b}_{jh}
\end{equation}
The matrix that shows the average net cage displacement for each of the examined environmental conditions is:
\[
   \mathbf{\bar{M}} = 
   \begin{bmatrix}
     \bar{x}_{1,1}   & \dots & \bar{x}_{1,s} \\
     \bar{y}_{1,1}   & \dots & \bar{y}_{1,s} \\
     \bar{z}_{1,1}   & \dots & \bar{z}_{1,s} \\
      & \dots  \\
     \bar{x}_{321,1}  & \dots & \bar{x}_{321,s} \\
     \bar{y}_{321,1}  & \dots & \bar{y}_{321,s} \\
     \bar{z}_{321,1}  & \dots & \bar{z}_{321,s} \\
   \end{bmatrix}
\]

\subsection*{Machine learning functional relationships between current and PCA coefficients}
To develop the low-fidelity component of the NARG methodology, which predicts net cage deformation that corresponds to FhSim solution, we create a standard GP model to learn the functional relationship between current speed and the net cage deformation. Specifically, the GP model maps the current characteristics to the PCA coefficients, $\bar{b}_1$, $\bar{b}_2$, $\bar{b}_3$. Then, we utilize the inner sum of Equation \ref{eq:rec} 
($\sum_{j=1}^k \phi_j\cdot \bar{b}_j$) to reconstruct the net cage deformation for an individual environmental condition. In other words, we obtain the predicted $x_i$, $y_i$, $z_i$, where $i = 1, \ldots, 321$. Similar approaches—incorporating PCA for dimensionality reduction and the development of surrogate models using machine learning methods—have been widely applied across various disciplines due to their efficacy in feature extraction and data simplification \cite{champenois2024machine, rudy2022prediction, guth2022wave, katsidoniotaki2023reduced,  guth2024}. In our case, performing the dimension reduction step helps reduce the complexity of the data describing the net cage deformation, making it easier to use machine learning methods to create the mapping.

\subsection*{Graph convolutional networks for net cage deformation}
Graph neural networks (GNNs) are mathematical models that can learn functions over graphs and are a learning approach for building predictive models on graph-structured data. Graphs differ from regular data in that they have a structure that neural networks must respect. Among the various types of GNNs, the graph convolutional networks (GCNs) have emerged as the most broadly applied model. GCNs are innovative due to their ability generalize the operation of convolutional neural networks to graphs, enabling the network to learn from the graph's topology and node features to make predictions. \\

\noindent \textbf{Representing net cage graph.} The net cage of our application can be viewed as graph-structured data, discretized into 321 nodes.We build a GCN model for the net cage based on the work of Kipf and Welling  \cite{kipf2016semi}. The net cage can be viewed as a graph $G$ consisting of $N = 321$ nodes connected through the edges, where $v_i$ and $v_j$ represent nodes $i$ and $j$. The edge features are represented by the adjacency matrix $\mathbf{A} \in \mathbb{R}^{N \times N}$ which encodes connections between the $N$ nodes, represents the graph connectivity which is a square matrix where each element $A_{ij}$ specifies the presence or absence of an edge from node $i$ to node $j$ in the graph. In other words, a non-zero element $A_{ij}$ implies a connection from node $i$ to node $j$, and a zero indicates no direct connection. In our net cage application, the adjacent matrix $\mathbf{A}$ captures the node connection shown in Figure \ref{fig:FhSim-Sensors} (b); the net cage consists of 10 layers, each with 32 nodes arranged in a circular pattern. Nodes within each layer are interconnected to form a circle, and each node is vertically connected to the node directly below it in the subsequent layer. This creates a consistent connection pattern between layers. Additionally, the final node (node 321) is connected to all nodes in the 10th layer, providing a central connection point. 
The $X \in \mathbb{R}^{L \times F}$ is the node feature matrix with $F$ being the number of the feature assigned to each node.
In our case, the $\mathbf{X}$ contains the features for each node which in our case are the $x_o$, $y_o$, $z_o$ coordinates of each node at rest and the sea conditions (i.e. current speed and direction). The graph is supposed to predict the matrix $Y \in \mathbb{R}^{L \times O}$, where $O$ is the number of output features per node. The matrix $\mathbf{Y}$ gives the $x$, $y$, $z$ coordinates of each node under the interaction of the cage with the sea condition.\\

\noindent \textbf{Message aggregation and update.} The GCN layers are based on the formulation by Kipf and Welling (2017) \cite{kipf2016semi}. At each layer, the graph convolution takes both the adjacency matrix \textbf{A} and the node features from the previous layer $\mathbf{H}^{l} \in \mathbb{R}^{L \times F_l}$ and outputs the node features for the next layer $\mathbf{H}^{l+1} \in \mathbb{R}^{L \times F_{l+1}}$, with $F_l$ and $F_{l+1}$ are the node features dimensions for layers $l$ and $l+1$, while $\mathbf{H}^0 = \mathbf{X}$ the input node feature matrix, following the formulation:
\begin{equation}
\label{eq:GCN}
\mathbf{H}^{l+1} = \sigma (\Tilde{\mathbf{D}}^{-0.5} \Tilde{\mathbf{A}} \Tilde{\mathbf{D}}^{-0.5} \mathbf{H}^l \mathbf{W}^l )          
\end{equation}
where $\sigma(\cdot)$ is the activation function, $\tilde{\mathbf{A}} = \mathbf{A} + \mathbf{I}_L$ is the adjacency matrix with added self-connections represented by the identity matrix $\mathbf{I}_L \in \mathbb{R}^{L \times L}$, $\tilde{\mathbf{D}}$ is the diagonal degree matrix of $\tilde{\mathbf{A}}$, and $\mathbf{W}^{l} \in \mathbb{R}^{F_l \times F_{l+1}}$ is a trainable weight matrix for layer $l + 1$. 
In our application, the data passes through three graph convolution layers. Each layer updates the node features by aggregating information from its neighbor nodes according to the graph structure. The final output represents the predicted node features which are the predicted displacements ($x$, $y$, $z$) for each node. Having several layers enables to further refine the node features by aggregating information from the graph structure and learning mode complex patterns.  The deformation of net-cage nodes is a highly non-linear process influenced, the ability of Swish activation function to introduce non-linearities is particularly beneficial, which is defined as:
\begin{equation}
    \sigma(x) = x \cdot \frac{1}{1 + e^{-x}}
\end{equation}
\noindent \textbf{Model training.} A notable point in our model is the implementation of a custom loss function that assigns different weights to the errors of each node by emphasizing regions of the net-cage that are more prone to displacement. At the node weights initialization stage nodes within layers 6 to 10, which are observed to undergo significant deformation, are assigned increased weights, with the highest weight allocated to the 321st node. The loss function computes the mean squared error (MSE) for each nodes prediction:
\begin{equation}
    \text{Loss} = \frac{1}{N} \sum_{i=1}^N w_i \cdot \left( \frac{1}{F} \sum_{j=1}^{F} (\hat{y}_{ij} - y_{ij})^2 \right)
\end{equation}
given $\hat{\mathbf{Y}} = [\hat{\mathbf{y}}_1, \hat{\mathbf{y}}_2, \ldots, \hat{\mathbf{y}}_N]$: Predicted displacements (matrix of shape $N \times F$), $\mathbf{Y} = [\mathbf{y}_1, \mathbf{y}_2, \ldots, \mathbf{y}_N]$: True displacements (matrix of shape $N \times F$),     $\mathbf{w} = [w_1, w_2, \ldots, w_N]$: Node weights (vector of length $N$). For the training process we use backpropagation to compute gradients and gradient descent optimiser to update the parameters for each layer.

\section*{Data availability statement} 
Code will become available to a dedicated depository after the paper is accepted. Correspondence and requests for code materials should be addressed to Eirini Katsidoniotaki or Themistoklis P. Sapsis.  The data that support the findings of this study are available from SINTEF Ocean AS but restrictions apply to the availability of these data, which were used under license for the current study, and so are not publicly available. Data are however available from the authors upon reasonable request and with permission of SINTEF Ocean AS.



\begin{thebibliography}{50}
\providecommand{\natexlab}[1]{#1}
\providecommand{\url}[1]{\texttt{#1}}
\expandafter\ifx\csname urlstyle\endcsname\relax
  \providecommand{\doi}[1]{doi: #1}\else
  \providecommand{\doi}{doi: \begingroup \urlstyle{rm}\Url}\fi

\bibitem[{ Johan Bergenas}()]{FP}
{ Johan Bergenas}.
\newblock {The Other Global Food Crisis}.
\newblock
  \url{https://foreignpolicy.com/2023/09/19/global-food-crisis-fishing-blue-foods-conflict-water-ocean-climate-resources/}.

\bibitem[Amundsen et~al.(2024)Amundsen, Xanthidis, F{\o}re, Ohrem, and
  Kelasidi]{amundsen2024aquaculture}
Herman~B Amundsen, Marios Xanthidis, Martin F{\o}re, Sveinung~J Ohrem, and
  Eleni Kelasidi.
\newblock Aquaculture field robotics: Applications, lessons learned and future
  prospects.
\newblock \emph{arXiv preprint arXiv:2404.12995}, 2024.

\bibitem[Babaee et~al.(2020)Babaee, Bastidas, Defilippo, Chryssostomidis, and
  Karniadakis]{babaee2020multifidelity}
Hessam Babaee, C~Bastidas, Michael Defilippo, C~Chryssostomidis, and
  GE~Karniadakis.
\newblock A multifidelity framework and uncertainty quantification for sea
  surface temperature in the massachusetts and cape cod bays.
\newblock \emph{Earth and Space Science}, 7\penalty0 (2):\penalty0
  e2019EA000954, 2020.
\newblock \doi{https://doi.org/10.1029/2019EA000954}.

\bibitem[Bi et~al.(2020)Bi, Zhao, Sun, Zhang, Guo, Wang, and
  Dong]{bi2020efficient}
Chun-Wei Bi, Yun-Peng Zhao, Xiong-Xiong Sun, Yao Zhang, Zhi-Xing Guo, Bin Wang,
  and Guo-Hai Dong.
\newblock An efficient artificial neural network model to predict the
  structural failure of high-density polyethylene offshore net cages in typhoon
  waves.
\newblock \emph{Ocean Engineering}, 196:\penalty0 106793, 2020.
\newblock \doi{https://doi.org/10.1016/j.oceaneng.2019.106793}.

\bibitem[Bjelland et~al.(2015)Bjelland, F{\o}re, Lader, Kristiansen, Holmen,
  Fredheim, Gr{\o}tli, Fathi, Oppedal, Utne, et~al.]{bjelland2015exposed}
Hans~V Bjelland, Martin F{\o}re, P{\aa}l Lader, David Kristiansen, Ingunn~M
  Holmen, Arne Fredheim, Esten~I Gr{\o}tli, Dariusz~E Fathi, Frode Oppedal,
  Ingrid~B Utne, et~al.
\newblock Exposed aquaculture in norway.
\newblock In \emph{OCEANS 2015-MTS/IEEE Washington}, pages 1--10. IEEE, 2015.

\bibitem[Chakraborty et~al.(2021)Chakraborty, Adhikari, and
  Ganguli]{chakraborty2021role}
Souvik Chakraborty, Sondipon Adhikari, and Ranjan Ganguli.
\newblock The role of surrogate models in the development of digital twins of
  dynamic systems.
\newblock \emph{Applied Mathematical Modelling}, 90:\penalty0 662--681, 2021.
\newblock \doi{https://doi.org/10.1016/j.apm.2020.09.037}.

\bibitem[Champenois and Sapsis(2024)]{champenois2024machine}
Bianca Champenois and Themistoklis Sapsis.
\newblock Machine learning framework for the real-time reconstruction of
  regional 4d ocean temperature fields from historical reanalysis data and
  real-time satellite and buoy surface measurements.
\newblock \emph{Physica D: Nonlinear Phenomena}, 459:\penalty0 134026, 2024.
\newblock \doi{https://doi.org/10.1016/j.physd.2023.134026}.

\bibitem[Chetan et~al.(2021)Chetan, Yao, and Griffith]{chetan2021multi}
Mayank Chetan, Shulong Yao, and D~Todd Griffith.
\newblock Multi-fidelity digital twin structural model for a sub-scale downwind
  wind turbine rotor blade.
\newblock \emph{Wind Energy}, 24\penalty0 (12):\penalty0 1368--1387, 2021.
\newblock \doi{https://doi.org/10.1002/we.2636}.

\bibitem[Conti et~al.(2024)Conti, Guo, Manzoni, Frangi, Brunton, and
  Nathan~Kutz]{conti2024multi}
Paolo Conti, Mengwu Guo, Andrea Manzoni, Attilio Frangi, Steven~L Brunton, and
  J~Nathan~Kutz.
\newblock Multi-fidelity reduced-order surrogate modelling.
\newblock \emph{Proceedings of the Royal Society A}, 480\penalty0
  (2283):\penalty0 20230655, 2024.
\newblock \doi{https://doi.org/10.1098/rspa.2023.0655}.

\bibitem[Cutajar et~al.(2019)Cutajar, Pullin, Damianou, Lawrence, and
  Gonz{\'a}lez]{cutajar2019deep}
Kurt Cutajar, Mark Pullin, Andreas Damianou, Neil Lawrence, and Javier
  Gonz{\'a}lez.
\newblock Deep gaussian processes for multi-fidelity modeling.
\newblock \emph{arXiv preprint arXiv:1903.07320}, 2019.

\bibitem[Desai et~al.(2023)Desai, Navaneeth, Adhikari, and
  Chakraborty]{desai2023enhanced}
Aarya~Sheetal Desai, N~Navaneeth, Sondipon Adhikari, and Souvik Chakraborty.
\newblock Enhanced multi-fidelity modeling for digital twin and uncertainty
  quantification.
\newblock \emph{Probabilistic Engineering Mechanics}, 74:\penalty0 103525,
  2023.
\newblock \doi{https://doi.org/10.1016/j.probengmech.2023.103525}.

\bibitem[F{\o}re and Thorvaldsen(2021)]{fore2021causal}
Heidi~Moe F{\o}re and Trine Thorvaldsen.
\newblock Causal analysis of escape of atlantic salmon and rainbow trout from
  norwegian fish farms during 2010--2018.
\newblock \emph{Aquaculture}, 532:\penalty0 736002, 2021.
\newblock \doi{https://doi.org/10.1016/j.aquaculture.2020.736002}.

\bibitem[F{\o}re et~al.(2024)F{\o}re, Alver, Alfredsen, Rasheed, Hukkel{\aa}s,
  Bjelland, Su, Ohrem, Kelasidi, Norton, et~al.]{fore2024digital}
Martin F{\o}re, Morten~Omholt Alver, Jo~Arve Alfredsen, Adil Rasheed, Thor
  Hukkel{\aa}s, Hans~V Bjelland, Biao Su, Sveinung~J Ohrem, Eleni Kelasidi,
  Tomas Norton, et~al.
\newblock Digital twins in intensive aquaculture—challenges, opportunities
  and future prospects.
\newblock \emph{Computers and Electronics in Agriculture}, 218:\penalty0
  108676, 2024.
\newblock \doi{https://doi.org/10.1016/j.compag.2024.108676}.

\bibitem[Gephart et~al.(2021)Gephart, Henriksson, Parker, Shepon, Gorospe,
  Bergman, Eshel, Golden, Halpern, Hornborg, et~al.]{gephart2021environmental}
Jessica~A Gephart, Patrik~JG Henriksson, Robert~WR Parker, Alon Shepon,
  Kelvin~D Gorospe, Kristina Bergman, Gidon Eshel, Christopher~D Golden,
  Benjamin~S Halpern, Sara Hornborg, et~al.
\newblock Environmental performance of blue foods.
\newblock \emph{Nature}, 597\penalty0 (7876):\penalty0 360--365, 2021.

\bibitem[Guth and Sapsis(2022)]{guth2022wave}
S.~Guth and T.~P. Sapsis.
\newblock Wave episode based gaussian process regression for extreme event
  statistics in ship dynamics: Between the scylla of karhunen--lo{\`e}ve
  convergence and the charybdis of transient features.
\newblock \emph{Ocean Engineering}, 266:\penalty0 112633, 2022.

\bibitem[Guth et~al.(2024)Guth, Katsidoniotaki, and Sapsis]{guth2024}
S.~Guth, E.~Katsidoniotaki, and T.~P. Sapsis.
\newblock Statistical modeling of fully nonlinear hydrodynamic loads on
  offshore wind turbine monopile foundations using wave episodes and targeted
  cfd simulations through active sampling.
\newblock \emph{Wind Energy}, 27\penalty0 (1):\penalty0 75--100, 2024.

\bibitem[Han et~al.(2022)Han, Leem, Choi, and Chung]{han2022drives}
Kiuk Han, Kyounghee Leem, Young~Rok Choi, and Keunsuk Chung.
\newblock What drives a country’s fish consumption? market growth phase and
  the causal relations among fish consumption, production and income growth.
\newblock \emph{Fisheries Research}, 254:\penalty0 106435, 2022.
\newblock \doi{https://doi.org/10.1016/j.fishres.2022.106435}.

\bibitem[Huan et~al.(2020)Huan, Li, Li, and Chen]{huan2020prediction}
Juan Huan, Hui Li, Mingbao Li, and Bo~Chen.
\newblock Prediction of dissolved oxygen in aquaculture based on gradient
  boosting decision tree and long short-term memory network: A study of chang
  zhou fishery demonstration base, china.
\newblock \emph{Computers and Electronics in Agriculture}, 175:\penalty0
  105530, 2020.
\newblock \doi{https://doi.org/10.1016/j.compag.2020.105530}.

\bibitem[Huang et~al.(2006)Huang, Tang, and Liu]{huang2006dynamical}
Chai-Cheng Huang, Hung-Jie Tang, and Jin-Yuan Liu.
\newblock Dynamical analysis of net cage structures for marine aquaculture:
  Numerical simulation and model testing.
\newblock \emph{Aquacultural Engineering}, 35\penalty0 (3):\penalty0 258--270,
  2006.
\newblock \doi{https://doi.org/10.1016/j.aquaeng.2006.03.003}.

\bibitem[Katsidoniotaki et~al.(2023)Katsidoniotaki, Guth, G{\"o}teman, and
  Sapsis]{katsidoniotaki2023reduced}
Eirini Katsidoniotaki, Stephen Guth, Malin G{\"o}teman, and Themistoklis~P
  Sapsis.
\newblock Reduced order modeling of wave energy systems via sequential bayesian
  experimental design and machine learning.
\newblock 2023.

\bibitem[Katsidoniotaki et~al.(2024)Katsidoniotaki, Su, Kelasidi, and
  Sapsis]{katsidoniotaki2024}
Eirini Katsidoniotaki, Biao Su, Eleni Kelasidi, and Themistoklis Sapsis.
\newblock Integrating mac ine learning for real-time structural monitoring of
  net cages.
\newblock In \emph{34th International Ocean and Polar Engineering Conference
  (ISOPE 2024)}, 2024.

\bibitem[Kelasidi and Svendsen(2023)]{kelasidi2023robotics}
Eleni Kelasidi and Eirik Svendsen.
\newblock Robotics for sea-based fish farming.
\newblock In \emph{Encyclopedia of Smart Agriculture Technologies}, pages
  1--20. Springer, 2023.
\newblock \doi{https://doi.org/10.1007/978-3-031-24861-0_202}.

\bibitem[Kelasidi et~al.(2022)Kelasidi, Su, Caharija, F{\o}re, Pedersen, and
  Frank]{kelasidi2022autonomous}
Eleni Kelasidi, Biao Su, Walter Caharija, Martin F{\o}re, Magnus~Oshaug
  Pedersen, and Kevin Frank.
\newblock Autonomous monitoring and inspection operations with uuvs in fish
  farms.
\newblock \emph{IFAC-PapersOnLine}, 55\penalty0 (31):\penalty0 401--408, 2022.
\newblock \doi{https://doi.org/10.1016/j.ifacol.2022.10.461}.

\bibitem[Kennedy and O'Hagan(2000)]{kennedy2000predicting}
Marc~C Kennedy and Anthony O'Hagan.
\newblock Predicting the output from a complex computer code when fast
  approximations are available.
\newblock \emph{Biometrika}, 87\penalty0 (1):\penalty0 1--13, 2000.
\newblock \doi{https://doi.org/10.1093/biomet/87.1.1}.

\bibitem[Kipf and Welling(2016)]{kipf2016semi}
Thomas~N Kipf and Max Welling.
\newblock Semi-supervised classification with graph convolutional networks.
\newblock \emph{arXiv preprint arXiv:1609.02907}, 2016.

\bibitem[Kristiansen and Faltinsen(2012)]{kristiansen2012modelling}
Trygve Kristiansen and Odd~M Faltinsen.
\newblock Modelling of current loads on aquaculture net cages.
\newblock \emph{Journal of Fluids and Structures}, 34:\penalty0 218--235, 2012.
\newblock \doi{https://doi.org/10.1016/j.jfluidstructs.2012.04.001}.

\bibitem[Lai et~al.(2023)Lai, Yang, He, Pang, Song, and Sun]{lai2023digital}
Xiaonan Lai, Liangliang Yang, Xiwang He, Yong Pang, Xueguan Song, and Wei Sun.
\newblock Digital twin-based structural health monitoring by combining
  measurement and computational data: An aircraft wing example.
\newblock \emph{Journal of Manufacturing Systems}, 69:\penalty0 76--90, 2023.
\newblock \doi{https://doi.org/10.1016/j.jmsy.2023.06.006}.

\bibitem[Li et~al.(2023)Li, Xu, Guo, and Yang]{li2023application}
Xiaofei Li, Langxing Xu, Hainan Guo, and Lu~Yang.
\newblock Application of graph convolutional neural networks combined with
  single-model decision-making fusion neural networks in structural damage
  recognition.
\newblock \emph{Sensors}, 23\penalty0 (23):\penalty0 9327, 2023.
\newblock \doi{https://doi.org/10.3390/s23239327}.

\bibitem[Liu et~al.(2023)Liu, Han, Wang, and Liu]{liu2023modelling}
Xinyu Liu, Xu~Han, Honghui Wang, and Guijie Liu.
\newblock A modelling and updating approach of digital twin based on surrogate
  model to rapidly evaluate product performance.
\newblock \emph{The International Journal of Advanced Manufacturing
  Technology}, 129\penalty0 (11):\penalty0 5059--5074, 2023.
\newblock \doi{https://doi.org/10.1007/s00170-023-12646-w}.

\bibitem[Martin et~al.(2021)Martin, Tsarau, and Bihs]{martin2021numerical}
Tobias Martin, Andrei Tsarau, and Hans Bihs.
\newblock A numerical framework for modelling the dynamics of open ocean
  aquaculture structures in viscous fluids.
\newblock \emph{Applied Ocean Research}, 106:\penalty0 102410, 2021.
\newblock \doi{https://doi.org/10.1016/j.apor.2020.102410}.

\bibitem[Meng and Karniadakis(2020)]{meng2020composite}
Xuhui Meng and George~Em Karniadakis.
\newblock A composite neural network that learns from multi-fidelity data:
  Application to function approximation and inverse pde problems.
\newblock \emph{Journal of Computational Physics}, 401:\penalty0 109020, 2020.
\newblock \doi{https://doi.org/10.1016/j.jcp.2019.109020}.

\bibitem[Meng et~al.(2021)Meng, Babaee, and Karniadakis]{meng2021multi}
Xuhui Meng, Hessam Babaee, and George~Em Karniadakis.
\newblock Multi-fidelity bayesian neural networks: Algorithms and applications.
\newblock \emph{Journal of Computational Physics}, 438:\penalty0 110361, 2021.
\newblock \doi{https://doi.org/10.1016/j.jcp.2021.110361}.

\bibitem[Perdikaris et~al.(2017)Perdikaris, Raissi, Damianou, Lawrence, and
  Karniadakis]{perdikaris2017nonlinear}
Paris Perdikaris, Maziar Raissi, Andreas Damianou, Neil~D Lawrence, and
  George~Em Karniadakis.
\newblock Nonlinear information fusion algorithms for data-efficient
  multi-fidelity modelling.
\newblock \emph{Proceedings of the Royal Society A: Mathematical, Physical and
  Engineering Sciences}, 473\penalty0 (2198):\penalty0 20160751, 2017.
\newblock \doi{https://doi.org/10.1098/rspa.2016.0751}.

\bibitem[Popov and Sandu(2022)]{popov2022multifidelity}
Andrey~A Popov and Adrian Sandu.
\newblock Multifidelity data assimilation for physical systems.
\newblock \emph{Data Assimilation for Atmospheric, Oceanic and Hydrologic
  Applications (Vol. IV)}, pages 43--67, 2022.
\newblock \doi{https://doi.org/10.1007/978-3-030-77722-7_2}.

\bibitem[Rasmussen et~al.(2006)Rasmussen, Williams,
  et~al.]{rasmussen2006gaussian}
Carl~Edward Rasmussen, Christopher~KI Williams, et~al.
\newblock \emph{Gaussian processes for machine learning}, volume~1.
\newblock Springer, 2006.

\bibitem[Reite et~al.(2014)Reite, F{\o}re, Aars{\ae}ther, Jensen, Rundtop,
  Kyllingstad, Endresen, Kristiansen, Johansen, and Fredheim]{reite2014fhsim}
Karl-Johan Reite, Martin F{\o}re, Karl~Gunnar Aars{\ae}ther, J{\o}rgen Jensen,
  Per Rundtop, Lars~T Kyllingstad, Per~Christian Endresen, David Kristiansen,
  Vegar Johansen, and Arne Fredheim.
\newblock Fhsim—time domain simulation of marine systems.
\newblock In \emph{International Conference on Offshore Mechanics and Arctic
  Engineering}, volume 45509, page V08AT06A014. American Society of Mechanical
  Engineers, 2014.

\bibitem[Rudy and Sapsis(2022)]{rudy2022prediction}
S.~H. Rudy and T.~P Sapsis.
\newblock Prediction of intermittent fluctuations from surface pressure
  measurements on a turbulent airfoil.
\newblock \emph{AIAA Journal}, 60\penalty0 (7):\penalty0 4174--4190, 2022.

\bibitem[Saad et~al.(2023)Saad, Su, and Bj{\o}rnson]{saad2023web}
Aya Saad, Biao Su, and Finn~Olav Bj{\o}rnson.
\newblock A web-based platform for efficient and robust simulation of
  aquaculture systems using integrated intelligent agents.
\newblock \emph{Procedia Computer Science}, 225:\penalty0 4560--4569, 2023.
\newblock \doi{https://doi.org/10.1016/j.procs.2023.10.454}.

\bibitem[Su et~al.(2019)Su, Reite, F{\o}re, Aars{\ae}ther, Alver, Endresen,
  Kristiansen, Haugen, Caharija, and Tsarau]{su2019multipurpose}
Biao Su, Karl-Johan Reite, Martin F{\o}re, Karl~Gunnar Aars{\ae}ther,
  Morten~Omholt Alver, Per~Christian Endresen, David Kristiansen, Joakim
  Haugen, Walter Caharija, and Andrei Tsarau.
\newblock A multipurpose framework for modelling and simulation of marine
  aquaculture systems.
\newblock In \emph{International conference on offshore mechanics and arctic
  engineering}, volume 58837, page V006T05A002. American Society of Mechanical
  Engineers, 2019.

\bibitem[Su et~al.(2021)Su, Kelasidi, Frank, Haugen, F{\o}re, and
  Pedersen]{su2021integrated}
Biao Su, Eleni Kelasidi, Kevin Frank, Joakim Haugen, Martin F{\o}re, and
  Magnus~Oshaug Pedersen.
\newblock An integrated approach for monitoring structural deformation of
  aquaculture net cages.
\newblock \emph{Ocean Engineering}, 219:\penalty0 108424, 2021.
\newblock \doi{https://doi.org/10.1016/j.oceaneng.2020.108424}.

\bibitem[Su et~al.(2023)Su, Bj{\o}rnson, Tsarau, Endresen, Ohrem, F{\o}re,
  Fagertun, Klebert, Kelasidi, and Bjelland]{su2023towards}
Biao Su, Finn~O Bj{\o}rnson, Andrei Tsarau, Per~C Endresen, Sveinung~J Ohrem,
  Martin F{\o}re, Jan~T Fagertun, Pascal Klebert, Eleni Kelasidi, and Hans~V
  Bjelland.
\newblock Towards a holistic digital twin solution for real-time monitoring of
  aquaculture net cage systems.
\newblock \emph{Marine Structures}, 91:\penalty0 103469, 2023.
\newblock \doi{https://doi.org/10.1016/j.marstruc.2023.103469}.

\bibitem[Turnbull et~al.(2005)Turnbull, Bell, Adams, Bron, and
  Huntingford]{turnbull2005stocking}
James Turnbull, Alisdair Bell, Colin Adams, James Bron, and Felicity
  Huntingford.
\newblock Stocking density and welfare of cage farmed atlantic salmon:
  application of a multivariate analysis.
\newblock \emph{Aquaculture}, 243\penalty0 (1-4):\penalty0 121--132, 2005.
\newblock \doi{https://doi.org/10.1016/j.aquaculture.2004.09.022}.

\bibitem[Ubina et~al.(2023)Ubina, Lan, Cheng, Chang, Lin, Zhang, Lu, Cheng, and
  Hsieh]{ubina2023digital}
Naomi~A Ubina, Hsun-Yu Lan, Shyi-Chyi Cheng, Chin-Chun Chang, Shih-Syun Lin,
  Kai-Xiang Zhang, Hoang-Yang Lu, Chih-Yung Cheng, and Yi-Zeng Hsieh.
\newblock Digital twin-based intelligent fish farming with artificial
  intelligence internet of things (aiot).
\newblock \emph{Smart Agricultural Technology}, 5:\penalty0 100285, 2023.
\newblock \doi{https://doi.org/10.1016/j.atech.2023.100285}.

\bibitem[{United Nations}()]{UN}
{United Nations}.
\newblock {The World's Food Supply is Made Insecure by Climate Change}.
\newblock
  \url{https://www.un.org/en/academic-impact/worlds-food-supply-made-insecure-climate-change/}.

\bibitem[{World Economic Forum}(2022)]{WEF}
{World Economic Forum}.
\newblock Aquaculture: why the world needs a new wave of food production.
\newblock
  \url{https://www.weforum.org/agenda/2022/01/aquaculture-agriculture-food-systems/},
  2022.

\bibitem[{WWF}()]{WWF}
{WWF}.
\newblock Overfishing.
\newblock \url{https://www.worldwildlife.org/threats/overfishing/}.

\bibitem[Zhang et~al.(2022{\natexlab{a}})Zhang, Wu, Jiang, Choi, and
  Zhou]{zhang2022multi}
Lili Zhang, Yuda Wu, Ping Jiang, Seung-Kyum Choi, and Qi~Zhou.
\newblock A multi-fidelity surrogate modeling approach for incorporating
  multiple non-hierarchical low-fidelity data.
\newblock \emph{Advanced Engineering Informatics}, 51:\penalty0 101430,
  2022{\natexlab{a}}.
\newblock \doi{https://doi.org/10.1016/j.aei.2021.101430}.

\bibitem[Zhang et~al.(2022{\natexlab{b}})Zhang, Gui, Qu, and
  Feng]{zhang2022netting}
Ziliang Zhang, Fukun Gui, Xiaoyu Qu, and Dejun Feng.
\newblock Netting damage detection for marine aquaculture facilities based on
  improved mask r-cnn.
\newblock \emph{Journal of Marine Science and Engineering}, 10\penalty0
  (7):\penalty0 996, 2022{\natexlab{b}}.
\newblock \doi{https:// doi.org/10.3390/jmse10070996}.

\bibitem[Zhao et~al.(2019)Zhao, Bi, Sun, and Dong]{zhao2019prediction}
Yun-Peng Zhao, Chun-Wei Bi, Xiong-Xiong Sun, and Guo-Hai Dong.
\newblock A prediction on structural stress and deformation of fish cage in
  waves using machine-learning method.
\newblock \emph{Aquacultural Engineering}, 85:\penalty0 15--21, 2019.
\newblock \doi{https://doi.org/10.1016/j.aquaeng.2019.01.003}.

\bibitem[Zhao et~al.(2022)Zhao, Lian, Bi, and Xu]{zhao2022digital}
Yun-Peng Zhao, Likai Lian, Chun-Wei Bi, and Zhijing Xu.
\newblock Digital twin for rapid damage detection of a fixed net panel in the
  sea.
\newblock \emph{Computers and Electronics in Agriculture}, 200:\penalty0
  107247, 2022.
\newblock \doi{https://doi.org/10.1016/j.compag.2022.107247}.

\end{thebibliography}

\section*{Acknowledgements}
This work was supported by the Knut and Alice Wallenberg Foundation Postdoctoral Scholarship Program at MIT - KAW 2022.0334,  the Office of Naval Research (Grant no N00014-21-1-2357), by RACE internal funding (SINTEF Ocean AS) and by the Research Council of Norway (RCN) project: CHANGE (no. 313737). The experimental data utilized in this paper for validation of the framework were obtained from RACE internal funding and FLEXAQUA (no. 284361) project funded from RCN.

\section*{Author contributions statement}
All authors conceived the idea. E.Kat. and T.P.S developed the methodology for the multifidelity digital twin framework. E.Kat. analysed the data, developed the code and implemented the framework. B.S. conducted the FhSim simulations. B.S. and E.Kel. provided the field sensor measurements. E.Kat. and T.P.S. analysed the results. B.S. and E.Kel. suggested applications for the digital twin's practical implementation. E.Kat. wrote the original manuscript, with all authors contributing to editing and reviewing the manuscript.

\section*{Additional information}
\noindent \textbf{Competing interests statement}\\
\noindent The authors declare no competing interests.\\

\end{document}